\documentclass[draft]{agujournal2019}
\usepackage{url} 
\usepackage{lineno}
\usepackage{soul}
\usepackage{csquotes}
\usepackage{siunitx}
\usepackage{enumitem}
\usepackage{amssymb}
\usepackage{subcaption}
\draftfalse

\begin{document}

\title{Finetuning a Weather Foundation Model with Lightweight Decoders for Unseen Physical Processes}

\authors{Fanny Lehmann\affil{1,2}, 
    Firat Ozdemir\affil{3},
    Benedikt Soja\affil{4},
    Torsten Hoefler\affil{5},
    Siddhartha Mishra\affil{1},
    Sebastian Schemm\affil{6}}

\affiliation{1}{Seminar for Applied Mathematics, ETH Zurich, Switzerland}
\affiliation{2}{ETH AI Center, ETH Zurich, Switzerland}
\affiliation{3}{Swiss Data Science Center, ETH Zurich, Switzerland}
\affiliation{4}{Institute of Geodesy and Photogrammetry, ETH Zurich, Switzerland}
\affiliation{5}{Scalable Parallel Computing Lab, ETH Zurich, Switzerland}
\affiliation{6}{Department of Applied Mathematics and Theoretical Physics, University of Cambridge, United Kingdom}

\correspondingauthor{Fanny Lehmann}{fanny.lehmann@ai.ethz.ch}

\vspace{2em}

\begin{abstract}
    Recent advances in AI weather forecasting have led to the emergence of so-called ``foundation models'', typically defined by expensive pretraining and minimal fine-tuning for downstream tasks. However, in the natural sciences, a desirable foundation model should also encode meaningful statistical relationships between the underlying physical variables. 

    This study evaluates the performance of the state-of-the-art Aurora foundation model in predicting hydrological variables, which were not considered during pretraining. We introduce a lightweight approach using shallow decoders trained on the latent representations of the pretrained model to predict these new variables. As a baseline, we compare this to fine-tuning the full model, which allows further optimization of the latent space while incorporating new variables into both inputs and outputs.

    The decoder-based approach requires 50\% less training time and 35\% less memory, while achieving strong accuracy across various hydrological variables and preserving desirable properties of the foundation model, such as autoregressive stability. Notably, decoder accuracy depends on the physical correlation between the new variables and those used during pretraining, indicating that Aurora's latent space captures meaningful physical relationships.
    
    In this sense, we argue that an important quality metric for foundation models in Earth sciences is their ability to be extended to new variables without a full fine-tuning. This provides a new perspective for making foundation models more accessible to communities with limited computational resources, while supporting broader adoption in Earth sciences.
\end{abstract}

\section{Introduction}
    With the recent developments in machine learning (ML), numerous large deep learning models have been proposed in weather and climate science. Among them, many studies claim to design \textquote{foundation models}, without necessarily defining which characteristics of a large pretrained deep neural network defines a foundation model. The rise of foundation models in weather and climate science follows the success of these models in Natural Language Processing (NLP) and Computer Vision (CV) where their definition is less complex than in natural science. For example, \citeA[p.3]{bommasani_opportunities_2021} defines a foundation model as \textquote{any model that is trained on broad data (generally using self-supervision at scale) that can be adapted (e.g., fine-tuned) to a wide range of downstream tasks}.

    The use of broad data is the most prevailing feature in weather and climate foundation models, starting from Climax \cite{nguyen_climax_2023} that builds on several CMIP6 datasets for pretraining while fine-tuning is performed on ERA5 reanalysis data. This multi-dataset approach was further adopted in ORBIT \cite{wang_orbit_2024} and Aurora \cite{bodnar_foundation_2025}, the latter exploring also additional fine-tuning tasks on atmospheric chemistry and ocean wave modelling. Since downstream tasks generally involve fine-tuning with new data and/or a new training objective, it is at this stage not clear if the expensive pretraining phase of a foundation model provides any additional benefit over a task-specific model. The latter would typically consist of less parameters and would be trained specifically on the new objective, e.g., high-resolution prediction \cite{adamov_building_2025, xu_artificial_2025}, terrestrial water storage prediction \cite{li_forecasting_2024, palazzoli_graice_2025}. 

    In contrast to NLP and CV, self-supervision is seldomly used when pretraining weather foundation models and the focus is typically on forecasting. Nevertheless, some models such as AtmoRep \cite{lessig_atmorep_2023} and Prithvi~WxC \cite{schmude_prithvi_2024} define a self-supervised pretraining task with masking in parallel or in complement to forecasting. Masking is a technique used during the pre-training phase with the aim to reconstruct the randomly masked regions (sometimes masking is also applied across variables). One should also mention WeatherGFM \cite{zhao_weathergfm_2024} that proposes a \textquote{generalist foundation model} to handle several weather understanding tasks by unifying the visual representation of different data sources and tasks. Although applied to satellite data (which prevents comparison with the above-mentioned models), WeatherGFM can realise several downstream tasks with few-shot learning, e.g., super-resolution, weather forecasting, image translation from satellite to radar image.

    However, the generic definition of a foundation model pays less attention to the underlying connection between physical variables governed by partial differential equations (PDEs). Although the training data are presented as images, derived from numerical simulations, observations, or a combination of both, they actually represent the values of physical variables, such as temperature and pressure, at specific times. Therefore, we argue that a foundation model in weather science should also capture underlying physical relationships between the predicted variables, thanks to an extensive training on substantial datasets. This point of view can be related to recent developments in designing general purpose foundation models for Physics governed by PDEs \cite{herde_poseidon_2024}, where it has been shown that pretraining on carefully selected sets of PDEs enables the models to learn rich latent physical representations and provides considerable gains for unseen downstream tasks, over task-specific models. 

    To provide further insights, this work examines how the latent space, which contains a representation of the multivariate distribution of the training data, can be employed to predict new variables exhibiting possible correlations with those originally included in the training data. This enables us to assess whether the statistical relationship between physical variables has been encoded. \citeA{bodnar_foundation_2025} demonstrate that the Aurora model can predict new variables by incorporating them as additional inputs and outputs, followed by fine-tuning the entire model for this specific downstream task. In the present work, we instead keep the Aurora model frozen in its pretrained state and learn new variables directly from Aurora's latent space representation. Following the afore-described conception of a foundation model, the latent space should indeed encode the physical relationships between weather variables, allowing it to transfer physical information to new but related physical processes.

    Our approach additionally addresses one limitation of current deep learning weather models, which is their restriction to a limited set of variables (e.g., temperature, pressure, humidity, wind velocity) while many adjacent research and application domains would need more variables for their applications. \citeA{hardy_leveraging_2025} showed that a Convolutional Neural Network (CNN) can be used to predict 100m wind velocity from the outputs of the AIFS model (which include 10m wind velocity). Since it uses the output variables, the CNN learns the visual correlations between the different variables and may miss the physical information encoded in the large pretrained latent space. The present work focuses on hydrological variables related to the water cycle (e.g., evaporation, runoff, soil moisture, terrestrial water storage), a field that has not yet been studied with weather foundation models. By learning the new variables directly by decoding the latent space, our approach provides a lightweight extension that alleviates the costs of fine-tuning the entire foundation model.

    In the following, Section \ref{sec:methods} details the Aurora model, the benchmark models, and the evaluation metrics. Section \ref{sec:data} describes the datasets used, while Section \ref{sec:results} presents the prediction results. Finally, Section \ref{sec:discussion} discusses the impacts and limitations of our method before drawing some conclusions in Section \ref{sec:conclusion}.

\section{Methods}
\label{sec:methods}
    All models considered in this work focus on forecasting. They consider inputs at current time $t$ (and previous time $t - \Delta t$, depending on the model) and predict outputs at the next time step $t + \Delta t$, where $\Delta t$ equals 6~hours if not specified otherwise.

\subsection{Aurora Baseline Model}
\begin{figure}[h]
        \centering
        \includegraphics[width=\linewidth]{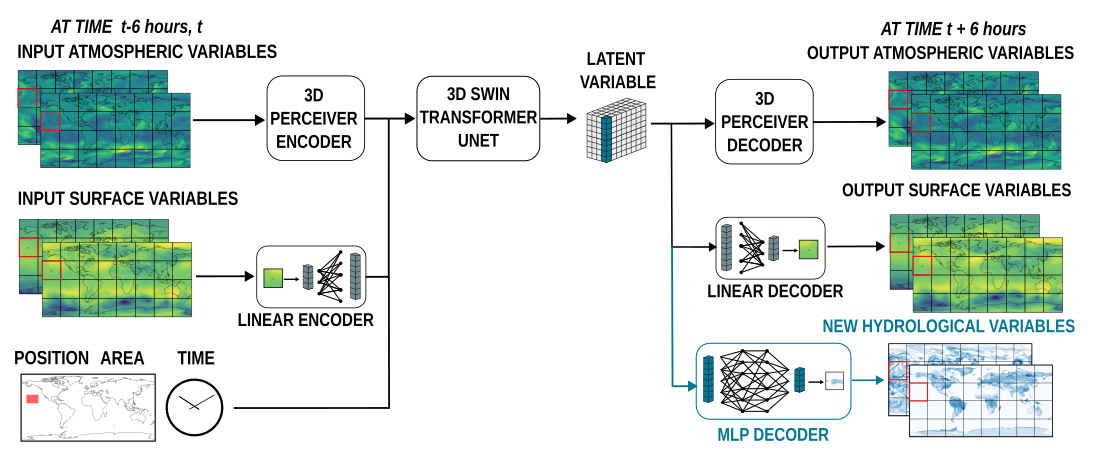}
        \caption{In black, the original Aurora model with its encoder-processor-decoder architecture. These blocks are frozen in our implementation. In blue, the additional Multi-Layer Perceptron (MLP) decoders that are trained to predict new physical variables.}
        \label{fig:aurora_decoders}
    \end{figure}

    The overall architecture of Aurora \cite{bodnar_foundation_2025} follows an encoder-processor-decoder structure (Fig.~\ref{fig:aurora_decoders}). The encoder takes as input images of size $T \times H \times W$ (with $T$=2 time steps) that can be static variables, surface variables, and atmospheric variables (in which case one image is given per pressure level). Similar to Vision Transformers (ViT), images are split into patches of size $T \times P \times P$. Surface and static variables are concatenated as image channels $V_s$. Similarly, atmospheric variables at each pressure level are concatenated as image channels $V_a$.
    The patches are then mapped to embedding vectors of dimension $E$ (with $E$=512) via a linear layer $T \times P \times P \times V_* \mapsto E$ (where $V_*$ is either $V_s$ or $V_a$). The final part of the encoder is a 3D Perceiver \cite{jaegle_perceiver_2021} that reduces the pressure levels of the atmospheric variables to a smaller number of 3 latent pressure levels.

    The processor is the core component and is responsible for learning the temporal dynamics. It is based on a 3D Swin Transformer U-Net and the reader is referred to the original work for more details on the architecture. The output of the processor is what we call the latent variable. Its size is $\frac{HW}{P^2} \times 4 \times 2E$, where the first dimension corresponds to the number of patches reshaped along a single dimension, the second dimension is the pressure dimension (3 latent pressure levels + 1 surface level), and the third dimension is the embedding.

    Symmetrically to the encoder, surface variables are decoded from first latent pressure level using a linear decoder. The remaining three latent pressure levels are expanded to the number of atmospheric pressure levels with a 3D Perceiver and subsequently decoded into the atmospheric variables with a linear decoder. The decoders of \cite{bodnar_foundation_2025} consist of a single linear layer per variable with $2E$ input neurons and $P^2$ output neurons, thus transforming the embeddings into a patch. 
    
    In our configuration, we use height $H$=720, width $W$=1440, patch size $P$=4, and embedding dimension $E$=512. This corresponds to the pretrained checkpoint available from \citeA{bodnar_foundation_2025}.  

\subsection{Decoder}
    To predict new variables, we keep the Aurora model frozen and train one additional decoder per new physical variable. Since we consider only surface variables, we choose a Multi-Layer Perceptron (MLP) with three layers of size $E$, $E/2$, $E/2$, and ReLU activations in-between. Since the Aurora model is frozen, we marginally increase the representative complexity of the decoders with additional non-linearities compared to the original 1-layer surface decoders. Each MLP decoder contains around \num{300 000} learnable parameters. In the following, the term \textit{decoders} means that only the MLP decoders are trained.

    New hydrological variables include ones that are defined only over land. 
    Accordingly, the loss function for these variables is computed based on a land-sea mask to exclude pixels over the oceans. The loss function is the Mean Absolute Error (MAE) with latitude-dependent weighting. The choice of the MAE follows Aurora implementation and the latitude-dependent weighting is commonly used to account for the geometric distortion of the globe when projected onto a rectangular grid \cite{chen_fuxi_2023, nguyen_climax_2023}. Training consists of 10 epochs with a batch size of 8 on 32 GH200 GPUs. The learning rate follows a cosine decay from \num{5d-4} down to \num{5d-5} preceded by a linear warmup for 100 steps.

\subsection{Benchmark Models}
    To compare the results of the decoders with other deep learning models, state-of-the-art models that predict at least precipitation are selected. GraphCast is based on a Graph Neural Network \cite{lam_learning_2023}. FuXi builds on Swin Transformer and integrates cascade models for different lead times \cite{chen_fuxi_2023}. ACE2 relies on Spherical Fourier Neural Operators and adds physical constraints to target long-term stability on climatic time scales \cite{watt-meyer_ace2_2024}. ACE2 predicts more variables than GraphCast and FuXi, in particular variables related to energy fluxes that are used in Section \ref{sec:results_energy}. ACE2 predictions are available on a \ang{1}~$\times$~\ang{1} grid and are bilinearly interpolated to \ang{0.25}~$\times$~\ang{0.25} for comparison with our Decoder (Section~\ref{sec:results_energy}). GraphCast and FuXi predictions are taken from WeatherBench2 \cite{rasp_weatherbench_2024}. 
    
    Furthermore, Aurora is fine-tuned for the new physical variables using a standard approach, where new variables serve as input and output. The Aurora baseline architecture is modified only by adding a linear encoder and a linear decoder for each new surface variable. Then, the full \qty{1.3}{billion}-parameter model is finetuned on a latitude-weighted MAE loss function that includes all variables. In the loss function, new variables have a weight of 1, while the other variables retain the same weights as in the original Aurora implementation. This model is named Aurora\textsuperscript{+} in the following. Aurora\textsuperscript{+} is fine-tuned on the same dataset as the decoders, for 10 epochs with a batch size of 1 (per GPU), with a learning rate undergoing a linear warmup for 500 steps up to \num{5e-4} and then a cosine decay.
    
    For the reference numerical predictions, we use the mean of the 50 ensembles from the ECMWF Integrated Forecasting System (IFS), available in WeatherBench2. The IFS model is a previous version underlying the ERA5 reanalysis, which also assimilates observations.

\subsection{Evaluation Metrics}
\label{subsec:metrics}
    The output variable dimension at each time step is $H \times W \times C$ where $C$=1 for surface variables, otherwise $C$ corresponds to the number of atmospheric levels. The agreement between predicted variables $\hat{X}$ and the reference $X$ is evaluated with several metrics. Since this work focuses on surface variables, the metrics definition below omits the third dimension for the sake of simplicity. The MAE
	\begin{equation}
		\text{MAE}(\hat{X}, X) = \frac{1}{HW} \sum_{i=1}^H \sum_{j=1}^W |\hat{X}(i,j) - X(i,j)|
		\label{eq:mae}
	\end{equation}
	and Root Mean Square Error (RMSE)
	\begin{equation}
		\text{RMSE}(\hat{X}, X) = \sqrt{\frac{1}{HW} \sum_{i=1}^H \sum_{j=1}^W \left( \hat{X}(i,j) - X(i,j) \right)^2}
		\label{eq:rmse}
	\end{equation}
	evaluate the point-wise accuracy. The RMSE gives more penalisation to outliers (such as isolated pixels with significantly lower or higher values than the reference). The bias
	\begin{equation}
		\text{bias}(\hat{X}, X) = \frac{1}{HW} \sum_{i=1}^H \sum_{j=1}^W \left (\hat{X}(i,j) - X(i,j) \right)
		\label{eq:bias}
	\end{equation}
	is a signed metric that indicates whether predictions are generally underestimated (i.e., bias is negative) or overestimated (i.e., bias is positive). The sample Pearson correlation coefficient (PCC) 
	\begin{equation}
		\text{PCC}(\hat{X}, X) = \frac{\sum_{i=1}^H \sum_{j=1}^W \left(\hat{X}(i,j) - \overline{\hat{X}}\right) \left( X(i,j) - \overline{X} \right)}{\sqrt{\sum_{i=1}^H \sum_{j=1}^W \left( \hat{X}(i,j) - \overline{\hat{X}}\right)^2} \sqrt{\sum_{i=1}^H \sum_{j=1}^W \left( X(i,j) - \overline{X} \right)^2}}
		\label{eq:pcc}
	\end{equation}
	evaluates the linear correlation between the predictions and the reference. A correlation of 1 is perfect, 0 indicates no correlation, and negative values indicate opposite correlation. In Equation~\ref{eq:pcc}, $\overline{X}:=\frac{1}{HW} \sum_{i=1}^H \sum_{j=1}^W X(i,j)$ is the average reference, and similarly, $\overline{\hat{X}}:=\frac{1}{HW} \sum_{i=1}^H \sum_{j=1}^W \hat{X}(i,j)$ is the average prediction. 
	
    The Wasserstein-1 distance \cite{ramdas_wasserstein_2015} compares the distribution of predicted values to the reference distribution
    \begin{equation}
        W_1(\hat{X}, X) = \int_{-\infty}^{+\infty} |CDF_{\hat{X}}(x) - CDF_X(x)| dx 
        \label{eq:W1}
    \end{equation} where $CDF_X$ (resp. $CDF_{\hat{X}}$) is the empirical Cumulative Distribution Function (CDF) computed from samples $X_n$ (resp. $\hat{X}_n$). For variables with narrow distributions such as precipitation (with a majority of pixels being close to 0), the Wasserstein-1 distance is also computed in the log space, $W_1(\log(\hat{X}), \log(X))$.
	
	For precipitation specifically, two categorical metrics are introduced to capture the ability of the decoder to predict values above a given threshold. The Fraction Skill Score (FSS, \citeA{roberts_scale-selective_2008}) evaluates the fraction of pixels above a threshold $\alpha$ (with $\alpha$=\qtyproduct{1}{mm} and $\alpha$=\qtyproduct{5}{mm} in this work) in windows of size \qtyproduct{11 x 11}{pixels}. The FSS is defined as 
	\begin{equation}
		\text{FSS}(\hat{X}, X) = 1 - \frac{\sum_w (f_w(\hat{X}) - f_w(X))^2}{\sum_w f_w(\hat{X})^2 + \sum_w f_w(X)^2}
		\label{eq:fss}
	\end{equation} 
	where $f_w(X)$ (resp.\ $f_w(\hat{X})$) is the fraction of pixels above threshold $\alpha$ in the reference (resp.\ predicted) field and $w$ iterates over all windows in the $H \times W$ field. The Stable~Equitable~Error in Probability Space (SEEPS, \citeA{rodwell_new_2010}) classifies precipitation into three categories: `dry' when 6-hour precipitation is lower than \qty{0.25}{mm}, `light rain' between \qty{0.25}{mm} and a \textit{wet threshold} determined from climatology, and `heavy rain' above the \textit{wet threshold}.

\section{Data}
\label{sec:data}
	The main training dataset is the ERA5 reanalysis, widely used to train AI weather models thanks to its high quality and long time span \cite{hersbach_era5_2020}. Aurora uses five atmospheric variables on 13 pressure levels: temperature, specific humidity, geopotential, northward and eastward wind speed; four surface variables: 2-metre temperature, mean sea level pressure, northward and eastward 10-metre wind velocity; and three static variables: land-sea mask, geopotential at the surface, and the soil type. All these variables come from ERA5 and are given as inputs to Aurora to get the latent vector corresponding to each time step. 
	
	In addition, the training targets for the decoders include 6-hour accumulated potential evaporation, runoff (the sum of surface and subsurface runoff), and soil moisture (the sum of volumetric soil water layer weighted by the layer thickness for the first three layers defined in ERA5, i.e., down to 1-metre depth). While our main focus is on hydrological variables, we also add variables related to the surface and top of atmosphere energy balance: surface latent heat flux, surface sensible heat flux, surface net solar radiation, surface net thermal radiation, surface solar radiation downwards, surface thermal radiation downwards, top net solar radiation, and top net thermal radiation. All energy variables are 6-hour accumulations at T00, T06, T12, and T18 UTC, computed from the 1-hourly data given in ERA5 (the sum from T01 to T06 is chosen for the accumulation at T06, and similarly for T12, T18, T00).
	
	Due to known biases for precipitation in ERA5, the Multi-Source Weighted-Ensemble Precipitation version 2.80 (MSWEP) dataset is used instead \cite{beck_mswep_2019}. MSWEP is a combination of gauge-, satellite-, and reanalysis-based data providing global coverage of total precipitation at a \ang{0.1} spatial resolution, 3-hourly temporal resolution, from 1979 to 2022.  MSWEP was found to be more accurate than ERA5 for hydrological modelling \cite{gebrechorkos_global-scale_2024}. This work uses a bilinear spatial interpolation to regrid MSWEP to \ang{0.25} and sums two consecutive samples to obtain 6-hour accumulated precipitation. Precipitation is log-transformed as $\log(1 + x/\epsilon)$ where $\epsilon = 10^{-5}$, following \citeA{rasp_weatherbench_2020}. 
	
    The training dataset comprises \num{17 000} samples between May 2002 and June 2014, while the validation dataset comprises \num{1 700} samples between July 2014 and December 2015, and the test dataset contains \num{1460} samples between January 2020 and December 2020.

\section{Results}
\label{sec:results}
\subsection{Prediction of Hydrological Variables}
\begin{figure}[p]
        \centering
        \includegraphics[width=\linewidth]{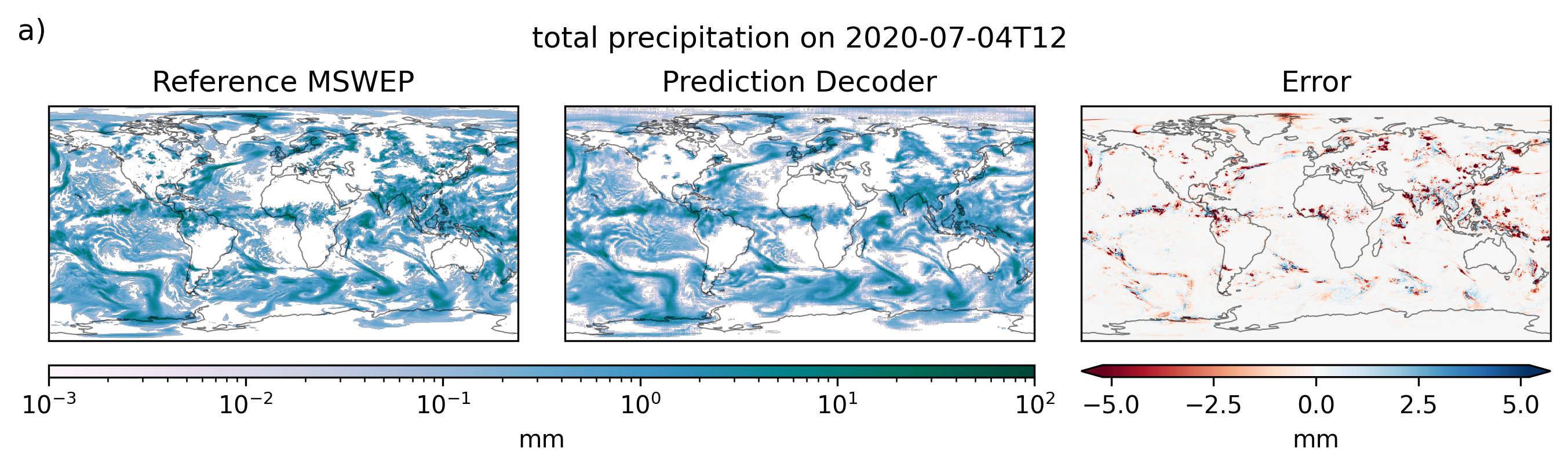}
        \includegraphics[width=\linewidth]{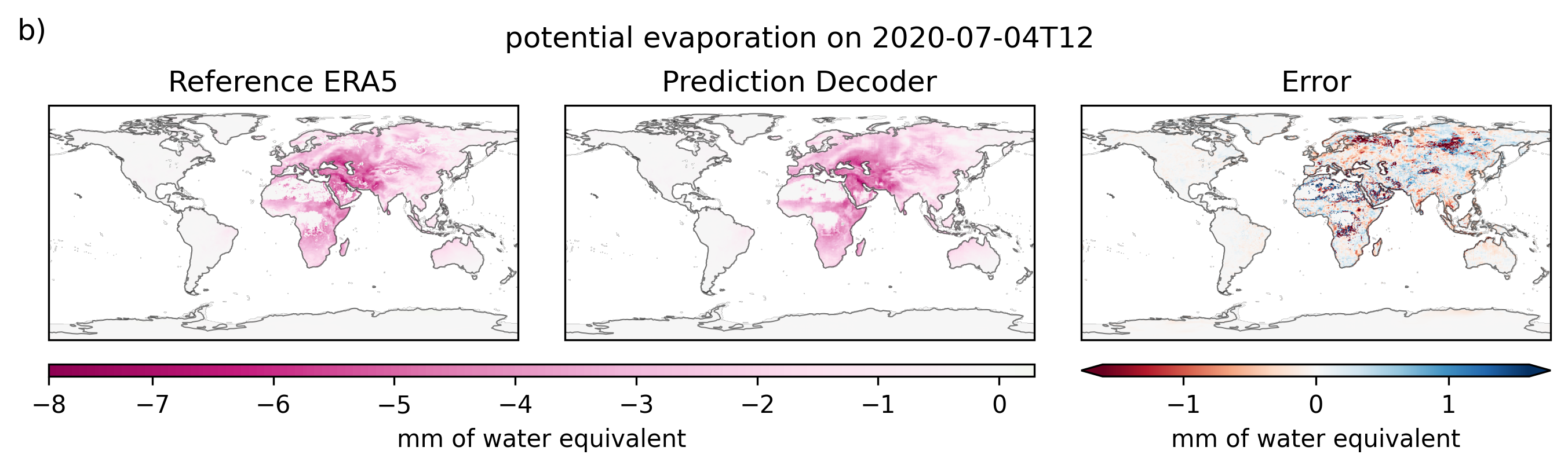}
        \includegraphics[width=\linewidth]{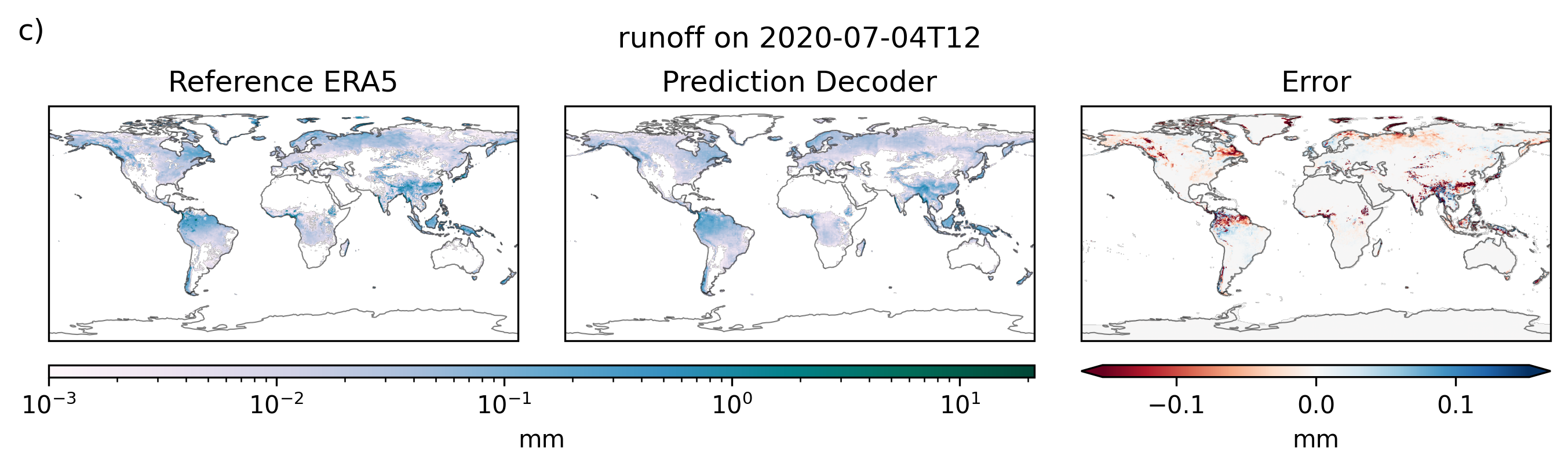}
        \includegraphics[width=\linewidth]{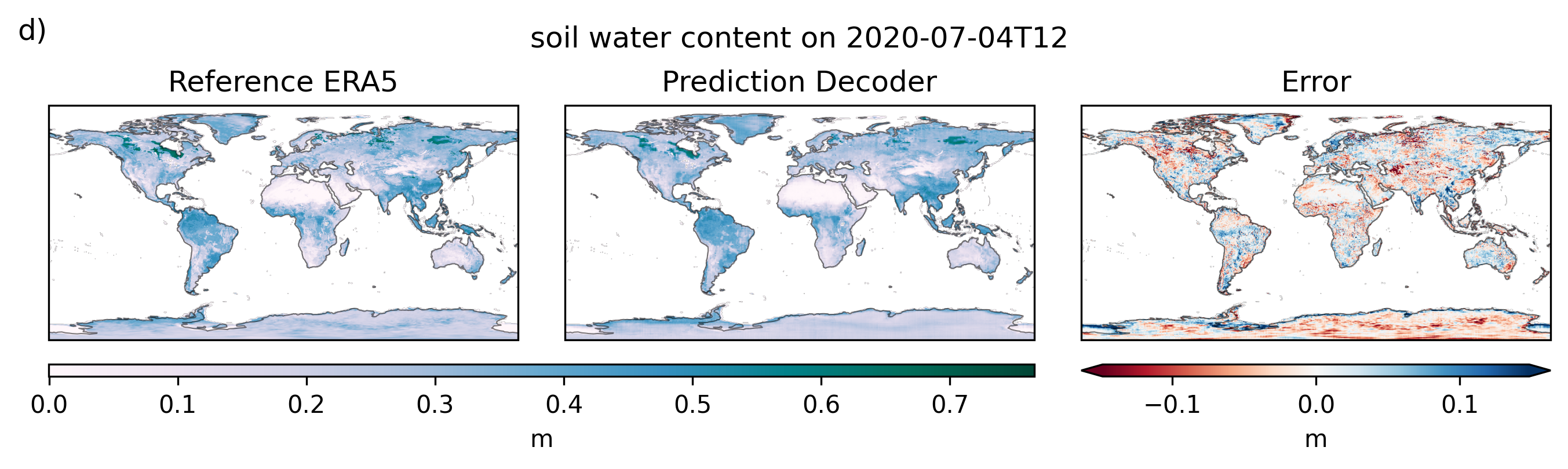}
        \caption{Next-step prediction at t+6h of a) total precipitation [mm/6hr], b) potential evaporation [mm/6hr], c) runoff [mm], d) soil moisture [m]. Left column is the reference, middle column is the prediction from the decoder, right column is the point-wise error between the prediction and the reference on July 4th, 2020.}
        \label{fig:maps_decoder}
    \end{figure}

    Figure~\ref{fig:maps_decoder} illustrates the predictions of four hydrological variables with the decoders on July 4th, 2020: total precipitation, potential evaporation, runoff, and soil moisture. First, a visual inspection of Fig. \ref{fig:maps_decoder} indicates that the decoder prediction is close to the baseline in many key locations. This is especially remarkable given that hydrological variables exhibit visual patterns very different from the original atmospheric variables (Fig.~\ref{fig:maps_original_aurora} vs.\ Fig.~\ref{fig:maps_decoder}). As expected from MAE-based training, regions of high amplitude or strong gradients present the greatest challenges for the decoders. Overall, the decoders, despite their simple architectures, capture the visual patterns of the new variables.
    
    Accuracy metrics are quantified for all 1460 test samples in 2020 in Tab.~\ref{tab:metrics_hydro} and Tab.~\ref{tab:metrics_precip}. Metrics confirm that, overall, decoders are able to predict all hydrological variables under study. Good accuracy is observed for potential evaporation and soil moisture, with correlation coefficients above \num{0.95} and relative MAE of \num{0.25} (resp.~\num{0.17}) for potential evaporation (resp.~soil moisture). The energy spectra confirm the good agreement at all spatial scales, with a slight underestimation by the decoders for small spatial scales (Fig.~\ref{fig:energy_spectra}). Runoff, however, shows higher errors (relative MAE of \num{0.66}) and low correlation coefficients (\num{0.42}). Precipitation is discussed in detail in the following Section~\ref{sec:results_precipitation}. 

    \begin{table}
    \caption{Metrics of our decoder (first column) and our Aurora\textsuperscript{+} predictions (second column) for the new hydrological variables. Predictions are done with a 6-hour lead time in 2020. Precipitation metrics are given in Tab.~\ref{tab:metrics_precip}. The metric definitions are given in Section \ref{subsec:metrics}. Better results are indicated in \textbf{bold}.}
    \label{tab:metrics_hydro}
    \centering
    \begin{tabular}{|ll|cc|}
        \hline
        Variable & Metric & Decoder & Aurora\textsuperscript{+} \\
        \hline
        potential evaporation & MAE (mm) & 0.040 & \textbf{0.019} \\
         & RMSE (mm) & 0.177 & \textbf{0.079} \\
         & bias (mm) & -0.003 & 0.003 \\
         & PCC & 0.958 & \textbf{0.992} \\
         & W1 & 0.007 & \textbf{0.005} \\
        \hline
        runoff & MAE (mm) & 0.005 & \textbf{0.003} \\
         & RMSE (mm) & 0.087 & \textbf{0.079} \\
         & bias (mm) & -0.003 & \textbf{-0.002} \\
         & PCC & 0.420 & \textbf{0.559} \\
         & W1 & 0.004 & \textbf{0.002} \\
        \hline
        soil moisture & MAE (m) & 0.015 & \textbf{0.001} \\
         & RMSE (m) & 0.034 & \textbf{0.003} \\
         & bias (m) & -0.002 & \textbf{0.001} \\
         & PCC & 0.969 & \textbf{0.999} \\
         & W1 & 3.181 & \textbf{0.578} \\
        \hline
        \end{tabular}
    \end{table}

    The accuracy obtained with Aurora\textsuperscript{+} is listed in Tab.~\ref{tab:metrics_hydro} to give an indication of the error that can be obtained when fine-tuning the full foundation model for new variables. However, we want to emphasise again that the goal is not to determine the best model between the decoders and Aurora\textsuperscript{+}, given the significant difference in required training effort (Tab.~\ref{tab:GPU_metrics}). The accuracy of the lightweight decoder closely matches the optimal Aurora\textsuperscript{+} for potential evaporation, while a larger difference is observed for soil moisture. This indicates that the latent space captured most of the physical information necessary for predicting potential evaporation, whereas soil moisture benefits from additional learning. This can be understood from the fact that potential evaporation is mostly determined by wind velocity, air temperature, and pressure, all variables included in the pre-training of Aurora. In contrast, soil moisture also depends on land surface processes absent from the original Aurora model, making it difficult to decode from the latent space using MLPs. 

    \begin{table}[h]
    \caption{Training speed evaluated by the number of Floating Point Operations Per Second (FLOPS) and the number of samples per second. Training memory requirements quantified with the allocated GPU memory.}
    \label{tab:GPU_metrics}
    \centering
    \begin{tabular}{|l|cc|}
        \hline
         & Decoder & Aurora\textsuperscript{+} \\
        \hline
        FLOPS & \num{4d11} & \num{31d12} \\
        samples/sec & 0.34 & 0.16 \\
        allocated GPU memory & 65GB & 99GB \\
        \hline
        \end{tabular}
    \end{table}

    Significant gains in training speed and GPU memory requirements compensate for the decreased accuracy of the decoders compared to Aurora\textsuperscript{+}. The decoders can process \num{0.34} samples per second, compared to \num{0.16} for Aurora\textsuperscript{+}, making their training twice faster (Tab.~\ref{tab:GPU_metrics}). The peak GPU memory necessary to train the decoders is also significantly reduced, from \qty{99}{GB} to \qty{65}{GB}.

\subsection{Precipitation Prediction and Comparison with Benchmarks}
\label{sec:results_precipitation}
    Precipitation is generally considered as a challenging variable to predict due to its high spatial and temporal variability. Nevertheless, precipitation maps in Fig.~\ref{fig:maps_decoder}a illustrate a good agreement between the prediction and the reference data. The regions affected by the largest errors can be roughly divided into two groups. Along the fronts of extratropical cyclones in extratropical latitudes, the predicted values are too low. In the tropics, the largest errors occur in regions of active convection along the equatorial belt. 
    
    The metrics in Tab.~\ref{tab:metrics_precip} show that the decoder achieves a good accuracy when predicting precipitation. FSS evaluates the ability to predict large precipitation values: \num{0.92} for a threshold of \qty{1}{mm} per \qty{6}{hour} and \num{0.82} for \qty{5}{mm} per \qty{6}{hour}. The correlation between the decoder prediction and the reference MSWEP dataset is satisfying (\num{0.71}).

    \begin{table}
    \caption{Precipitation metrics of our decoder (first column) and our Aurora\textsuperscript{+} predictions (second column). Predictions are done with a 6-hour lead time in 2020. The reference dataset is MSWEP. See Tab.~\ref{tab:metrics_precip_era5} for the same metrics with ERA5 as the reference dataset. The best metrics are indicated in \textbf{bold}, the second best are \underline{underlined}.}
    \label{tab:metrics_precip}
    \centering
    \begin{tabular}{|l|ccccc|}
        \hline
         & Decoder & Aurora\textsuperscript{+} & IFS & GraphCast & FuXi \\
        \hline
        MAE (mm) & \underline{0.32} & \textbf{0.22} & 0.44 & 0.38 & 0.36 \\
        RMSE (mm) & \underline{1.40} & \textbf{1.03} & 1.60 & 1.50 & 1.42 \\
        SEEPS & 0.35 & \textbf{0.23} & 0.34 & 0.33 & \underline{0.31} \\
        W1 & 0.246 & 0.17 & \textbf{0.06} & \underline{0.15} & 0.18 \\
        W1 log & 0.046 & \underline{0.034} & \textbf{0.029} & 0.047 & 0.042 \\
        bias (mm) & -0.14 & \underline{-0.096} & 0.10 & \textbf{-0.02} & \textbf{-0.02} \\
        FSS 1mm & \underline{0.92} & \textbf{0.97} & 0.89 & 0.90 & 0.91 \\
        FSS 5mm & \underline{0.82} & \textbf{0.93} & \underline{0.82} & 0.80 & 0.81 \\
        PCC & \underline{0.71} & \textbf{0.86} & 0.68 & 0.70 & \underline{0.71} \\
        \hline
        \end{tabular}
    \end{table}

    The fact that the decoder is skilful at predicting precipitation should be analysed in the light of the physical correlation between precipitation and atmospheric variables learnt by the original Aurora. In particular, it is known that regions of strong moisture flux convergence often coincide with precipitation and the convergence of moisture flux is the product of wind velocity and specific humidity. Therefore, despite the variability of precipitation and the visual difference with the original Aurora variables, the encoded statistical relationship between the atmospheric variables makes precipitation a learnable variable.

    As a reference, accuracy metrics on two benchmark AI models (GraphCast and FuXi) are also provided in Tab.~\ref{tab:metrics_precip}, alongside IFS numerical predictions. Results generally indicate better metrics for our decoder and Aurora\textsuperscript{+}. However, it should be noted that the reference dataset is MSWEP, which is not the training objective of GraphCast and FuXi. Setting ERA5 as the precipitation reference (while still training on MSWEP) degrades the metrics of the decoder and Aurora\textsuperscript{+} but keeps them comparable to IFS (Tab.~\ref{tab:metrics_precip_era5}). 

    \begin{figure}[p]
    \begin{subfigure}{\textwidth}
        \caption{}
        \includegraphics[width=\linewidth]{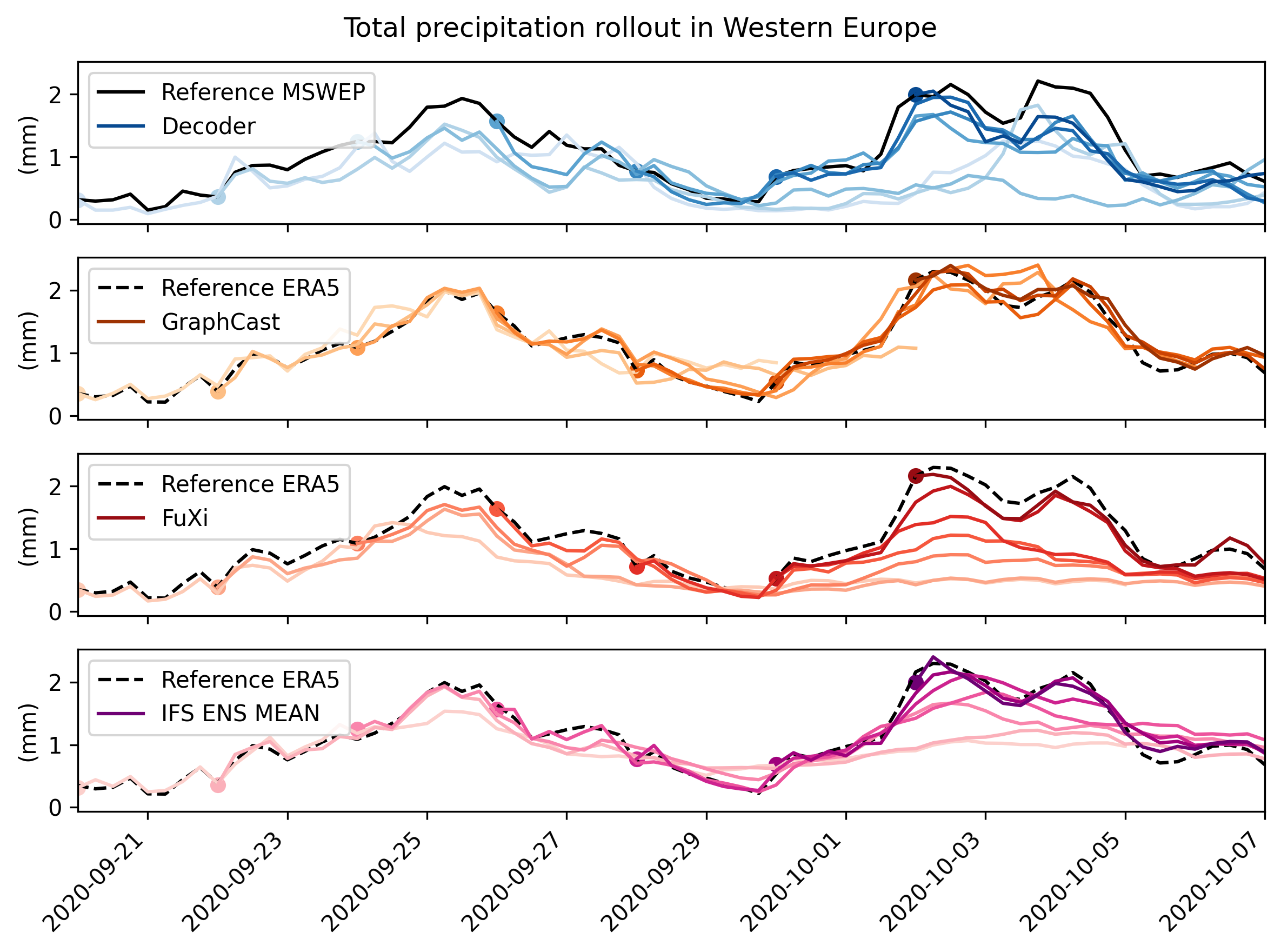}
    \end{subfigure}
    \hfill
    \begin{subfigure}{\textwidth}
        \caption{}
        \includegraphics[width=\linewidth]{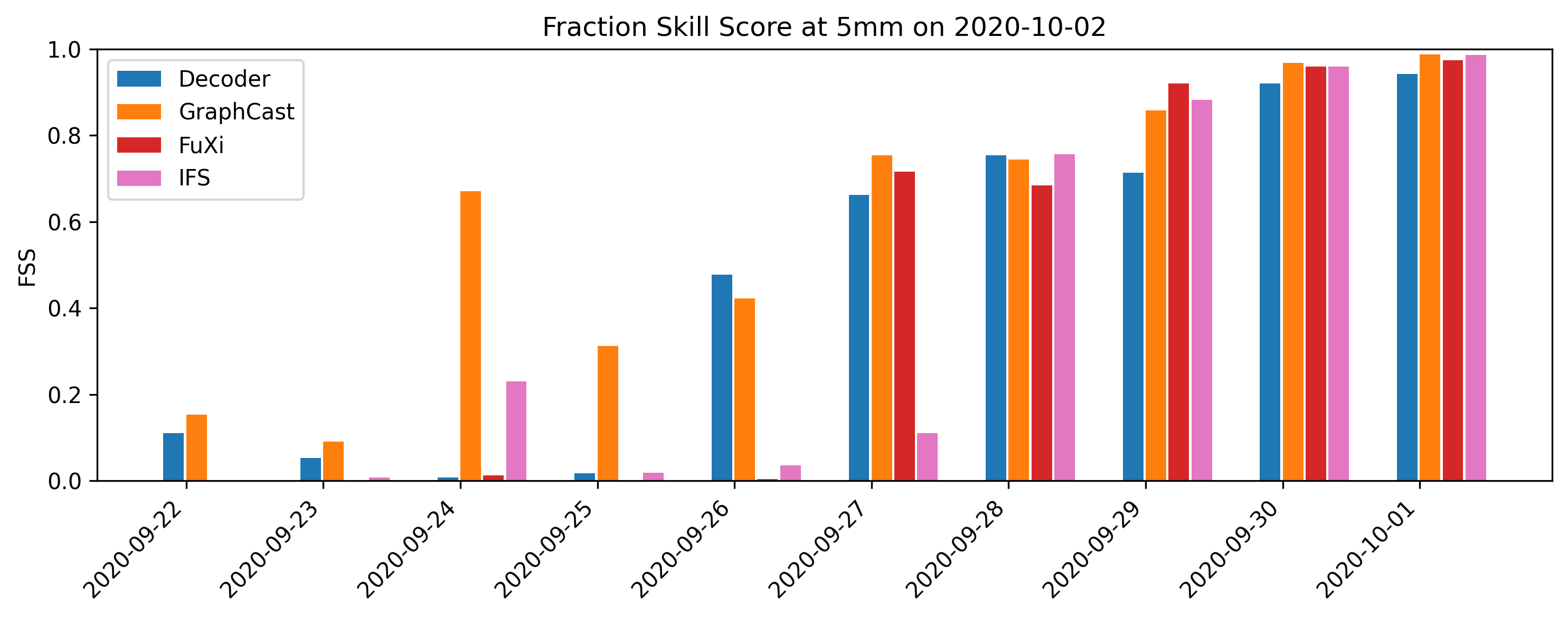}
    \end{subfigure}
        \caption{(a) Average of precipitation prediction in Western Europe (area shown in Fig.~\ref{fig:precip_europe_map_forecast}) for rollouts initialised every two days between 2020-09-20 and 2020-10-02 at T00. Each rollout is indicated with a different color shade (earlier rollouts with more pale colors) and the initial time is depicted with a dot. Each panel shows predictions with different models: our decoder (1st row), GraphCast (2nd row), FuXi (3rd row), IFS ensemble mean (4th row). (b) FSS with a 5mm threshold between observed and predicted precipitation on 2020-10-02T06. Predictions are obtained with rollouts initialized between 2020-09-22T00 and 2020-10-01T00. GraphCast, FuXi, and IFS are evaluated against ERA5 while our decoder is evaluated against MSWEP.}
        \label{fig:precip_europe_forecast}
    \end{figure}

    \begin{figure}[h]
        \centering
        \includegraphics[width=\linewidth]{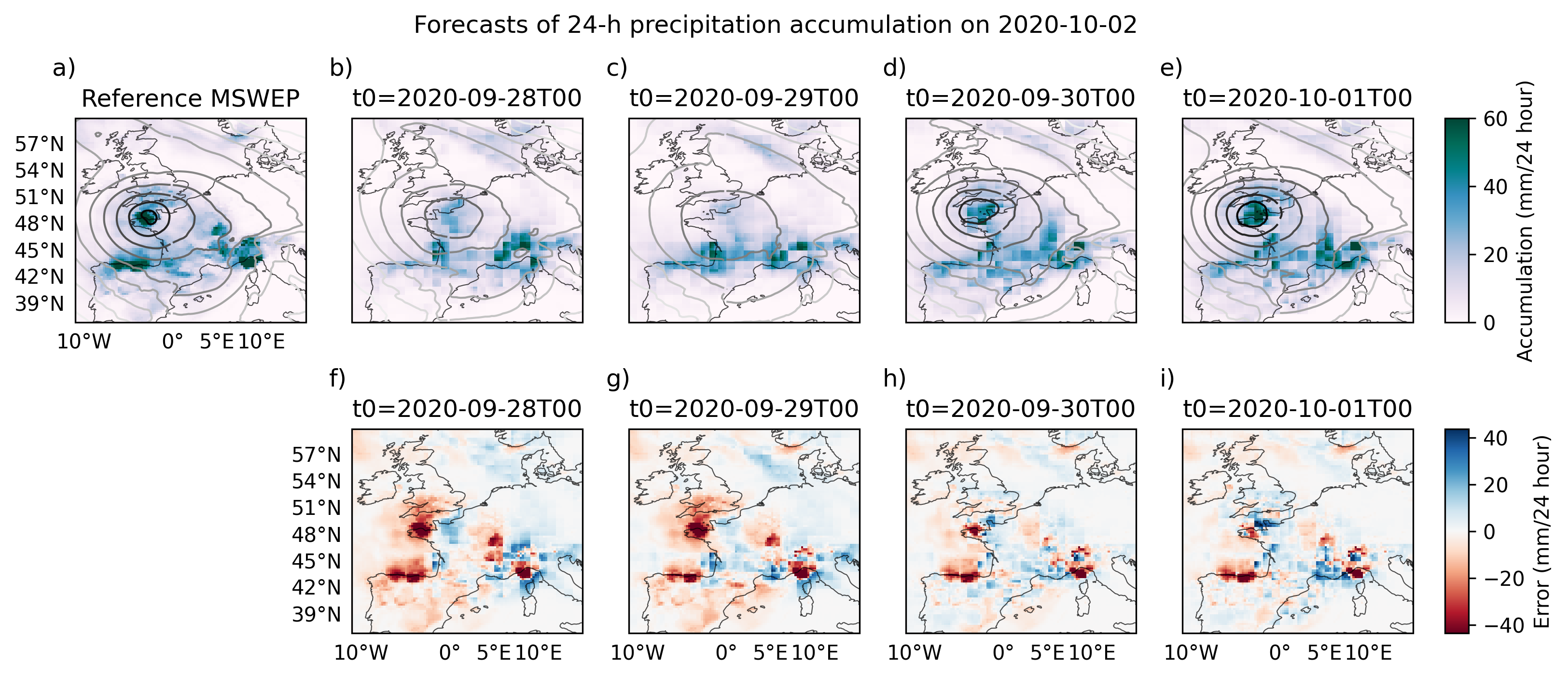}
        \caption{24-hour accumulation of precipitation in Western Europe on October 2nd, 2020, predicted by our decoder with a rollout from b) 2020-09-28 (18 time steps), c) 2020-09-29 (14 time steps), d) 2020-09-30 (10 time steps), e) 2020-10-01 (6 time steps). Panel a) shows the reference 24-hour accumulation on October 2nd, 2020. The bottom row (f-i) shows the error between the prediction and the reference for each rollout.}
        \label{fig:precip_europe_map_forecast}
    \end{figure}

    Next, consideration is given to the extreme precipitation associated with the extratropical cyclone \textquote{Alex} affecting Europe in early October 2020 (mainly Southern France and Northern Italy). Time series in Fig.~\ref{fig:precip_europe_forecast} indicate that the decoder is able to predict the large precipitation values averaged over the region (\ang{11}W, \ang{15}E, \ang{37}N, \ang{60}N) up to six days in advance. The longest lead time (18 6-hourly time steps) allows only anticipating the location of the three precipitation hot spots (Fig.~\ref{fig:precip_europe_map_forecast}b). The two regions in northern Spain and France, in particular, are too weak. Decreasing lead time improves the forecast location and intensity of intense precipitation, although the intensity is often too weak. The 6-hour accumulated precipitation observed on October 2nd at T00 was predicted by our decoder with a \qty{5}{mm} FSS of \num{0.94} (resp. \num{0.92}) one day (resp. two days) in advance. The underestimation of extreme events amplitude is in agreement with observations from various ML weather forecasting models \cite{liu_evaluation_2024}.

\subsection{Evaluation of Temporal Dynamics}
\label{sec:results_temporal}
\begin{figure}
        \centering
        \includegraphics[width=1\linewidth]{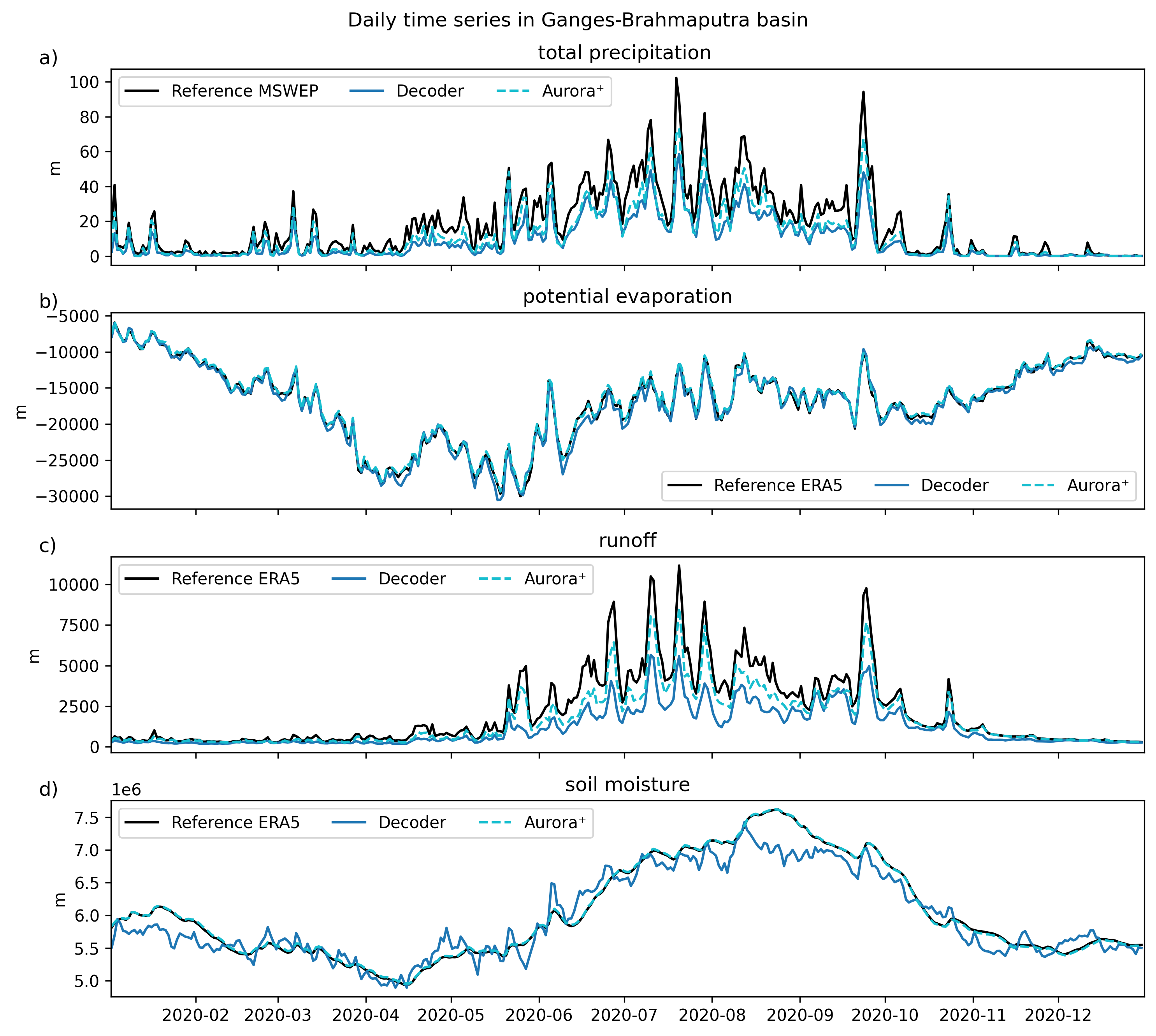}
        \caption{Daily average of 6-hour time series of a) total precipitation [m/6hr], b) potential evaporation [m of equivalent water/6hr], c) runoff [m], d) soil moisture [m] in the Ganges-Brahmaputra basin (shown in red in Fig.~\ref{fig:map_basins}). Time series are obtained as the weighted sum of pixels inside the basin.}
        \label{fig:hydro_basin}
    \end{figure}

\begin{figure}
        \centering
        \includegraphics[width=1\linewidth]{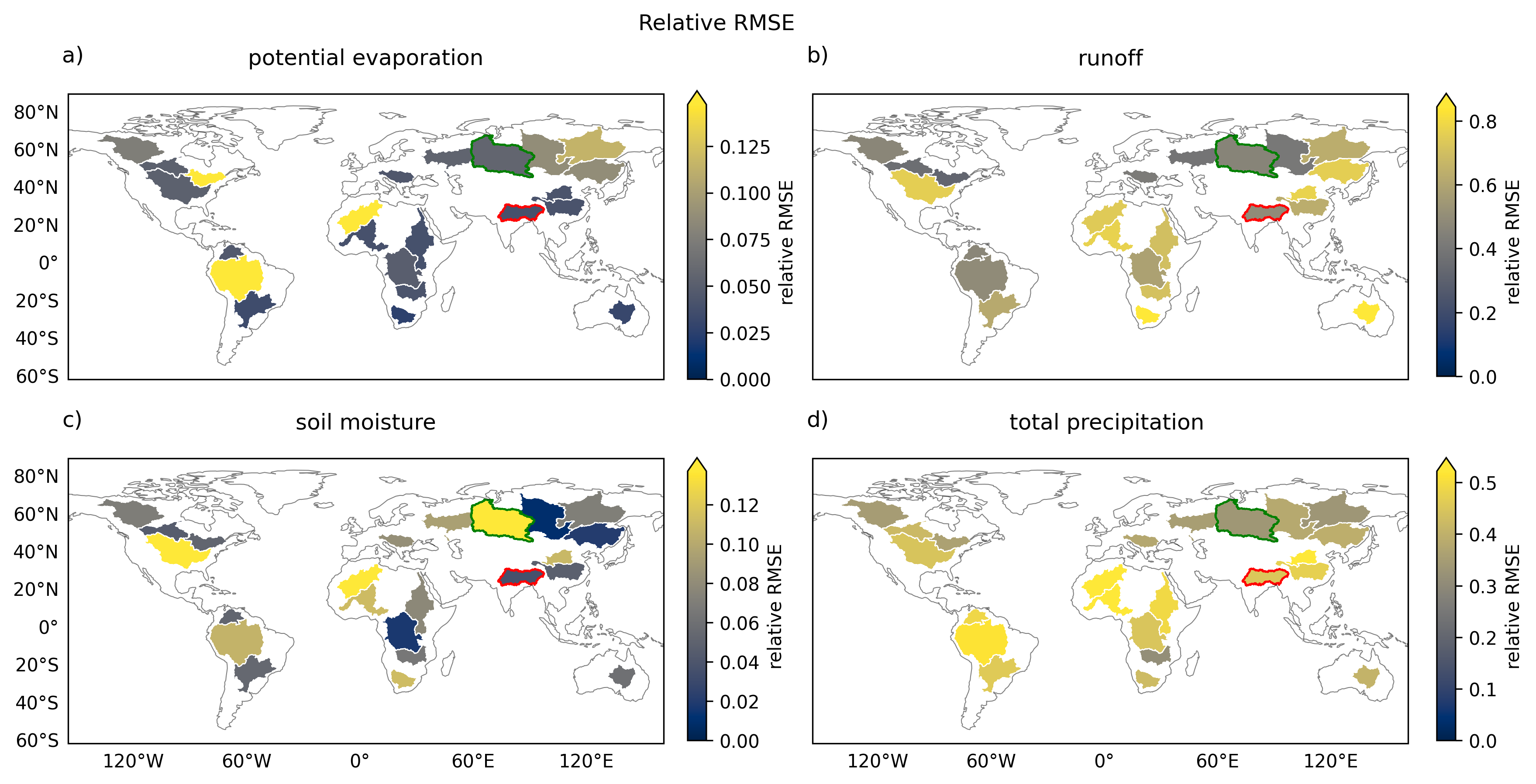}
        \caption{Relative RMSE computed on daily sums at the scale of a river basin. Highlighted in red is the Ganges-Brahmaputra basin, whose detailed time series are given in Fig.~\ref{fig:hydro_basin}, in green the Ob basin, with corresponding time series in Fig.~\ref{fig:hydro_basin_cold}.}
        \label{fig:map_basins}
    \end{figure}

    A spatial scale of interest in hydrology is the river basin (or catchment area), which is the area where water flows towards the river of the same name. Many hydrological studies aim at estimating hydrological variables from weather predictors, similar to the outcomes provided by our decoders. We computed hydrological time series in 24 major river basins\footnote{Amazon, Congo, Mississippi, Ob, Nile, Parana, Yenisey, Lena, West Sahara, Amur, Niger, Ganges-Brahmaputra, Yangtze, Mackenzie, Volga, Zambezi, Lake Eyre, Murray, Nelson, Saint Lawrence, Orinoco, Yellow river, Orange, Danube. Basin shapes were extracted from the level~3 of the HydroBasins dataset \cite{lehner_global_2013}.} (Fig.~\ref{fig:map_basins}) by summing the contribution of each pixel inside the basins (with weighting proportional to the latitude-dependent pixel area). Results in the Ganges-Brahmaputra basin are illustrated in Fig.~\ref{fig:hydro_basin}, with daily averages for an easier visualisation. 

    One can observe that the decoders reproduce the temporal dynamics of all hydrological variables for all seasons (Fig.~\ref{fig:hydro_basin}). As already noted from the metrics in Tab.~\ref{fig:maps_decoder}, potential evaporation shows excellent agreement with the reference, and the main temporal fluctuations of soil moisture are also correctly predicted by the decoders. Due to the slow soil moisture dynamics, an increment of 6~hours leads to very small changes. Nevertheless, despite training only with a 6-hour lead time, our decoder still captures the seasonal change. However, since the temporal dynamics of the soil moisture decoder is driven by the latent space, the decoder leads to soil moisture variations that are faster than the reference. In contrast, fine-tuning the entire foundation model correctly reproduces the slow temporal dynamics of soil moisture (Fig.~\ref{fig:hydro_basin}d).

    The overall evolution of runoff shows good agreement with the reference, but the decoder (and to a lesser extent, Aurora\textsuperscript{+}) underestimates the total runoff (Fig.~\ref{fig:hydro_basin}). The runoff peaks between June and October, correlated with high precipitation events, are especially noticeable. The decoder correctly identifies these peaks, but they are notably too weak compared to the reference and the fine-tuned Aurora\textsuperscript{+}. This reinforces the importance of the relationship between atmospheric and land surface processes for predicting the latter from the latent space. Runoff in the Ob basin, located in a cold climate, is largely driven by snow melt between April and September, a process the decoder cannot predict (Fig.~\ref{fig:hydro_basin_cold}). However, Aurora\textsuperscript{+} learns it, indicating that fine-tuning is superior in this case.

\subsection{Prediction of Energy Fluxes}
\label{sec:results_energy}
	Although the main focus of this work is on surface hydrological variables, decoders with the same architecture are also trained to predict a different set of variables, namely variables linked with energy fluxes. This allows studying the connection between the latent space and energy fluxes measured on the surface and at the top of the atmosphere. The decoders predict eight variables related to energy fluxes (in a training independent from the hydrological variables): surface latent/sensible heat flux, surface net solar/thermal radiation, surface solar/thermal radiation downwards, top net solar/thermal radiation (noting that the two latter top-of-atmosphere variables are not predicted by ACE2). 

    \begin{figure}[h]
        \centering
        \includegraphics[width=1\linewidth]{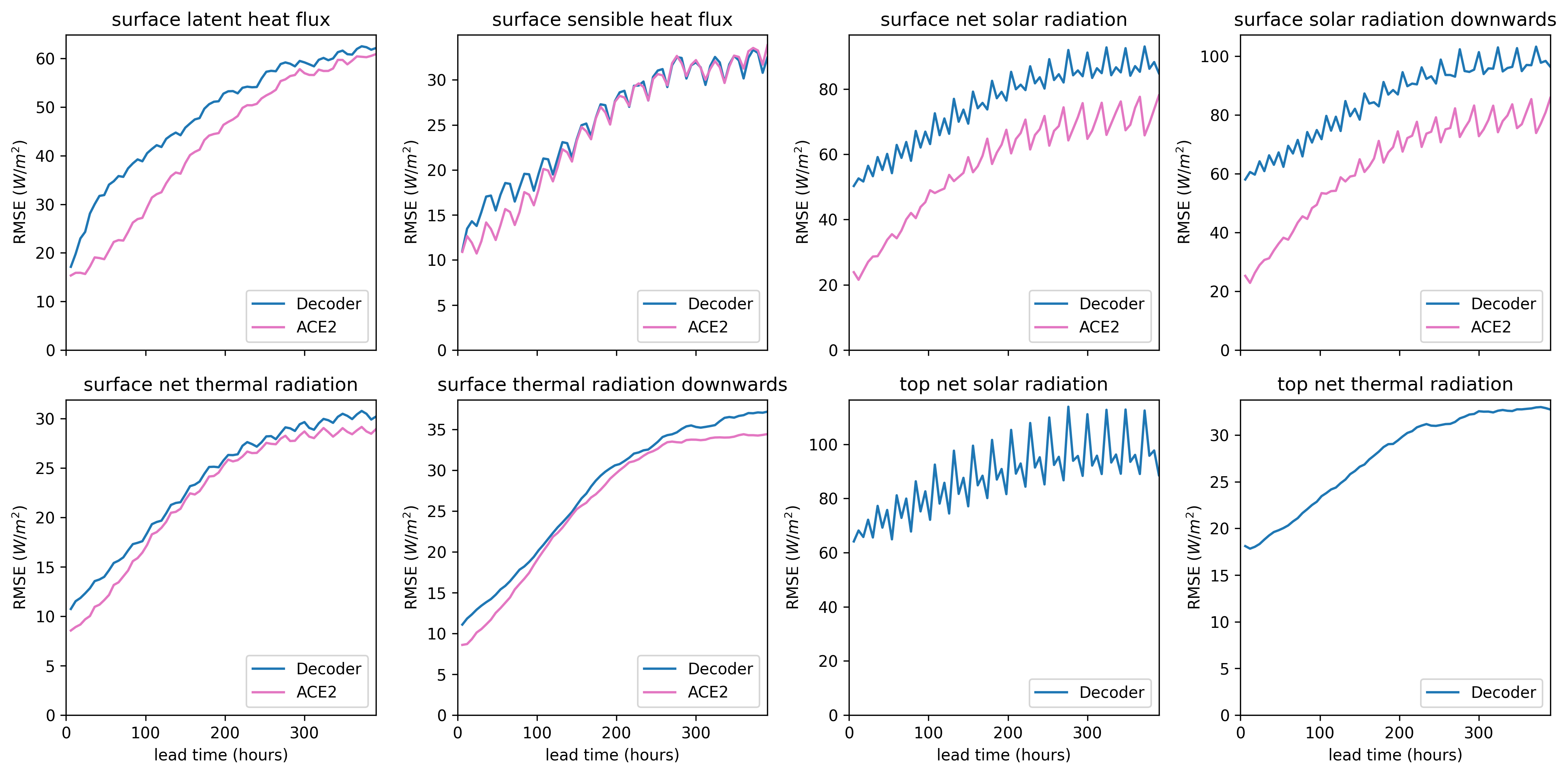}
        \caption{Root Mean Square Error (RMSE) of eight energy variables computed for lead times 6~hours to 384~hours (i.e., 16~days) with our decoder (blue) and ACE2 (pink). Errors should not be compared between variables as they take different values.}
        \label{fig:rollout_energy}
    \end{figure}

\begin{figure}[h]
    \begin{subfigure}{\textwidth}
        \caption{}
        \includegraphics[width=1\linewidth]{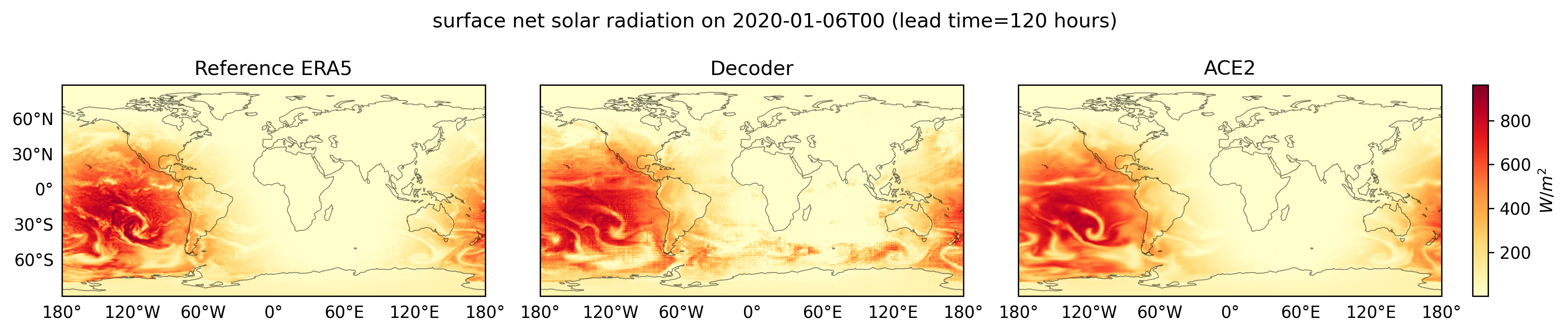}
    \end{subfigure}
    \hfill
    \begin{subfigure}{\textwidth}
        \caption{}
        \includegraphics[width=1\linewidth]{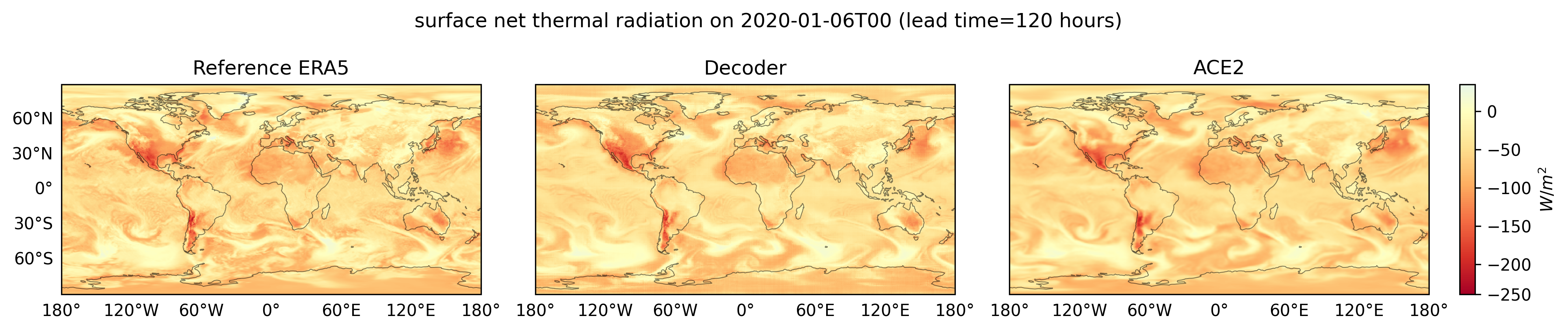}
    \end{subfigure}
        \caption{Map of a) surface net solar radiation and b) surface net thermal radiation (bottom panels) from the reference ERA5 (left), predicted by our decoder (middle), and predicted by ACE2 (right).}
        \label{fig:map_energy}
    \end{figure}

    Rollouts were initialised on the first day of each month from January 2020 to December 2020 and conducted for 384~hours (i.e., 16~days, 64 steps). Figures~\ref{fig:rollout_energy} and \ref{fig:map_energy} show that the decoders are able to predict all variables related to energy fluxes. The decoders yield RMSE close to those obtained with ACE2 for heat fluxes and thermal radiations, both for short and long lead times. Surface solar radiation presents higher errors with the decoder than ACE2, which can be understood by examining the maps in Fig.~\ref{fig:map_energy}. Indeed, solar fluxes concentrate on the region of the globe that receives the most solar radiation at that time of day, leaving half of the map without radiation (Fig.~\ref{fig:map_energy}a). These patterns differ from the original Aurora variables, whereas thermal radiation (Fig.~\ref{fig:map_energy}b) more closely resembles the type of fields already encoded in Aurora, making it easier to predict. ACE2 being fully trained with energy variables as target objectives yields accurate predictions for all energy variables. Figure~\ref{fig:rollout_energy} also shows that the decoder can learn variables at the top of the atmosphere (top net solar/thermal radiation), not only surface variables. 

\section{Discussion and Limitations}
\label{sec:discussion}
    New variables representing physical processes at the surface or at the top of the atmosphere can be learnt by the decoders. Despite their simple architecture, the decoders demonstrate skill across a wide range of variables, from hydrology to energy fluxes, and for different time scales. This shows that the latent space of a foundation model provides the appropriate input encoding to extend the model skills to new downstream tasks. This corroborates previous findings on the Poseidon PDE foundation model that was trained only on equations from fluid dynamics and could be fine-tuned on different physics (i.e., the reaction-diffusion Allen-Cahn equation) with a frozen latent space \cite{herde_poseidon_2024}. 
    
    The decoders also inherit the properties of the underlying foundation model, such as the auto-regressive stability. This allows the decoders to be readily applied for rollouts without the need to fine-tune them specifically for this objective. Such capability proved advantageous in anticipating the intense precipitation associated with the extratropical cyclone ``Alex''. Although the decoders depend solely on the rollout accuracy of Aurora, the rate at which prediction errors increase with longer lead times is similar to that of ACE2, a task-specific model fully trained with a rollout objective.

    Thanks to their lightweight architecture, training the decoders is twice as fast as fine-tuning the entire foundation model. During the backward pass, gradients are computed only for the decoders, leading to small computational graphs. This allows training without advanced checkpointing strategies, making them more accessible to a wide scientific community. The decoders can thus be considered as a \textit{frugal} extension of the foundation model.
    
    The accuracy of the decoders is influenced by the physical correlation between atmospheric processes learnt by the original Aurora, whose dynamics have therefore been encoded in the latent space. This was observed for precipitation, for instance, where the decoder achieves accurate predictions. In addition, the accuracy relates to the visual similarities between the original and new fields, as illustrated by the remaining artefacts on solar radiation (Fig.~\ref{fig:map_energy}a). The decoders cannot learn variables having little connection with atmospheric processes. This was observed, for instance, with terrestrial water storage, which encompasses all water bodies, from the surface to groundwater (Fig.~\ref{fig:maps_decoder_tws}).

    The decoders exhibit some limitations with high-resolution prediction, as can be seen in Fig.~\ref{fig:precip_europe_map}. The precipitation map shows some ``patchy'' artefacts with squares of \qtyproduct{4 x 4}{pixels}. These correspond to the patches created for the Swin transformer processor in the pretrained Aurora model. For variables with complex spatial variability, the decoders struggle to learn the correct values for each pixel inside the patch and resort to predictions at the patch scale. However, it is important to note that the predicted patches are not constant and the ``patchy'' artefacts are much less visible for smoother variables (e.g., soil moisture), as illustrated in Fig.~\ref{fig:maps_highres_japan}.

\section{Conclusion}
\label{sec:conclusion}
    This study investigates the predictions of new surface variables from the latent representations of a weather foundation model using only lightweight decoders and without fine-tuning the entire model. It can be concluded that new hydrological and energy variables can be predicted fairly accurately using these shallow MLP decoders. The decoders capture the spatial patterns of the new variables and their variations on different time scales, from sub-daily fluctuations to seasonal change. 

    The accuracy of the decoders improves when the target physical processes exhibit a close correlation with variables that have been encoded during pretraining. This indicates the effectiveness of the latent space of the Aurora foundation model in encoding statistical relationships between physical processes. 
    
    Although ERA5 is the main data source in this work, precipitation is taken from the MSWEP dataset to avoid known biases in ERA5 precipitation. The precipitation accuracy is satisfying with the decoders and the fine-tuned Aurora\textsuperscript{+} model, illustrating that different data sources can be used during training. This allows a selection of the most appropriate datasets from \textit{a priori} knowledge. 

    The decoders are competitive with benchmark models, although their accuracy does not exceed that of fully fine-tuned AI weather models. This is expected since only a small part of the overall model is trained. Furthermore, forecasts at time $t+\Delta t$ do not rely on target variable observations at time $t$, which are not available at input. On the other hand, their main benefit is the considerably reduced computational and memory costs. These advantages are potentially significant for startup companies and small research labs without access to substantial High-Performance Computing (HPC) systems.

    Our findings indicate that rich latent space representations allow accurate predictions of new variables with lightweight extensions. This calls for developing future foundation models that encompass a large number of physical processes, such that their latent space readily encodes correlations between diverse physical processes. Such foundation models would allow a broad adoption by diverse sectors with various needs and encourage a frugal use of AI resources.

\clearpage

\appendix
\section{Training Details}
    \begin{figure}[h]
        \centering
        \includegraphics[width=0.7\linewidth]{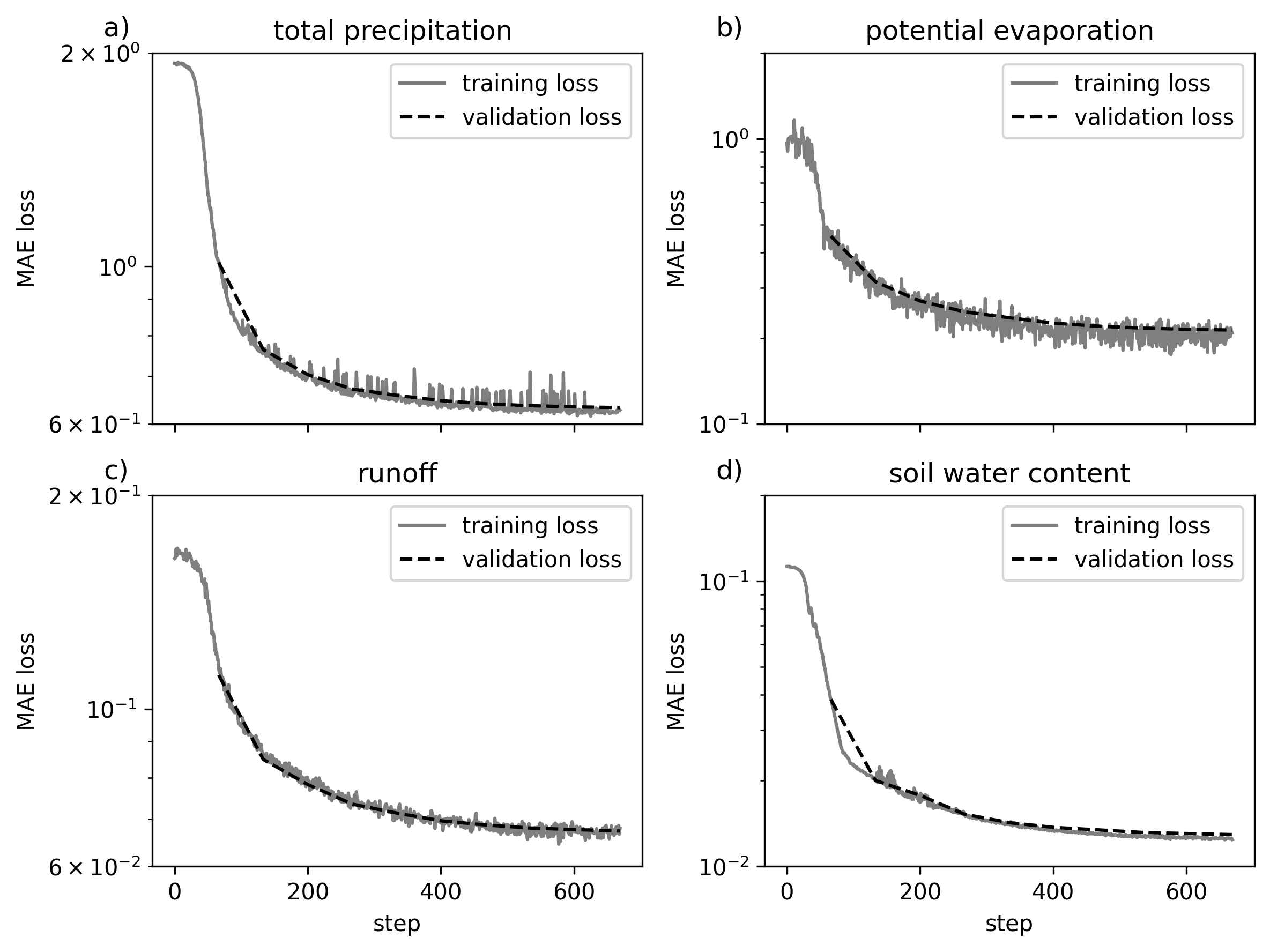}
        \caption{Training loss (grey line) and validation loss (dashed black line) for the decoders}
        \label{fig:loss_decoder}
    \end{figure}

    \begin{figure}[h]
        \centering
        \includegraphics[width=0.7\linewidth]{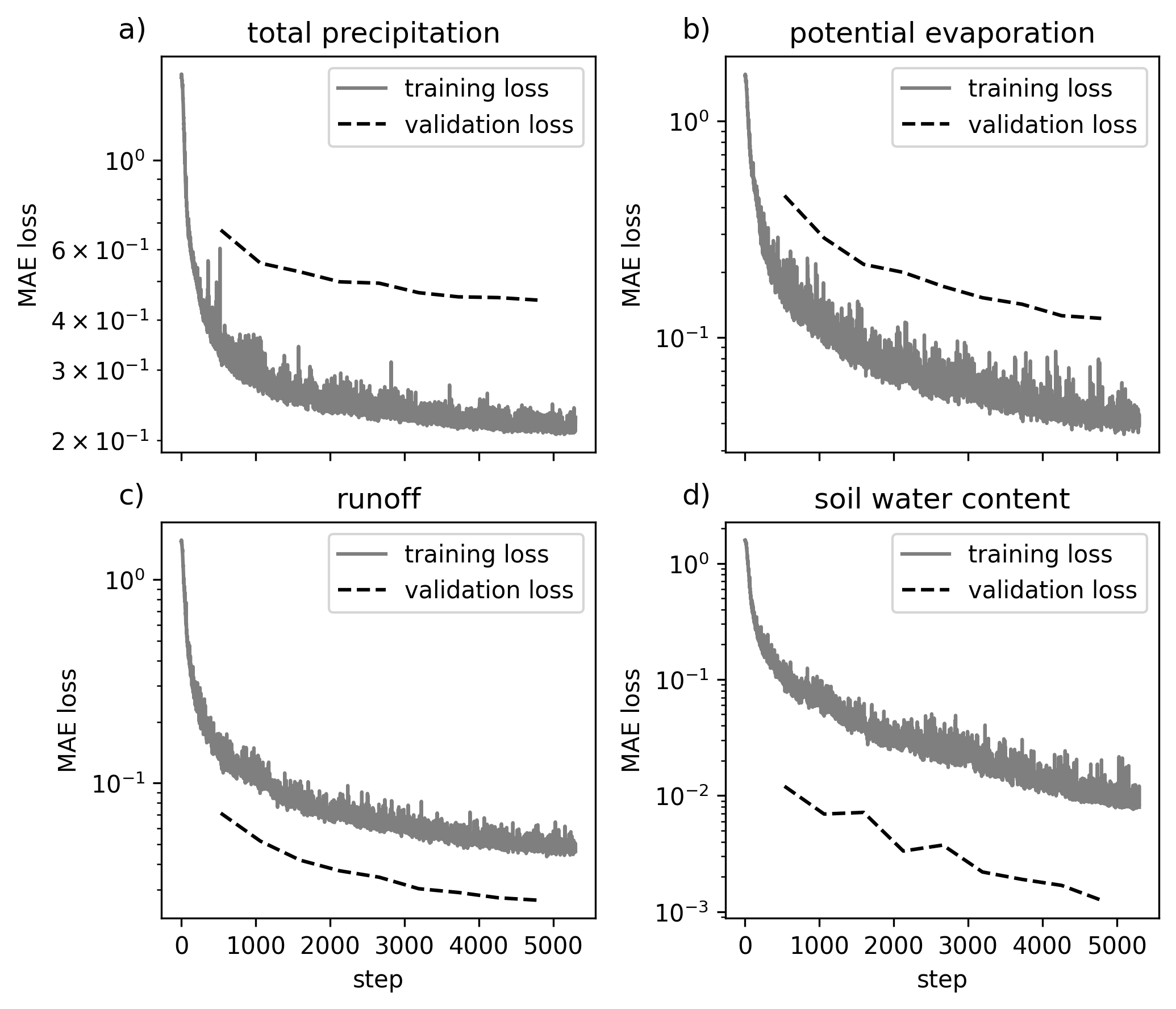}
        \caption{Training loss (grey line) and validation loss (dashed black line) for Aurora\textsuperscript{+}}
        \label{fig:loss_aurora}
    \end{figure}

\clearpage
\section{Prediction of Surface Variables}
\begin{figure}[p]
        \centering
        \includegraphics[width=\linewidth]{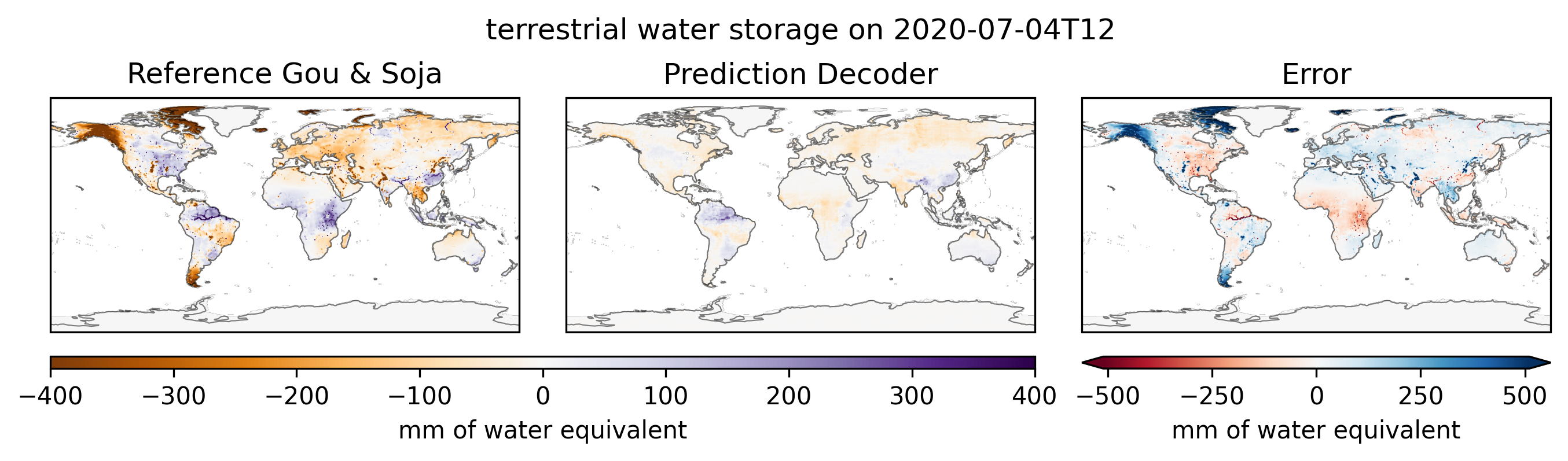}
        \caption{Next-step prediction at t+6h of terrestrial water storage. Left column is the reference, middle column is the prediction from the decoder, right column is the point-wise error between the prediction and the reference on July 4th, 2020.}
        \label{fig:maps_decoder_tws}
    \end{figure}

	In Fig.~\ref{fig:maps_decoder_tws}, terrestrial water storage is taken from \cite{gou_global_2024} and linearly interpolated to 6-hour time steps.
	
    \begin{figure}[h]
        \centering
        \includegraphics[width=0.9\linewidth]{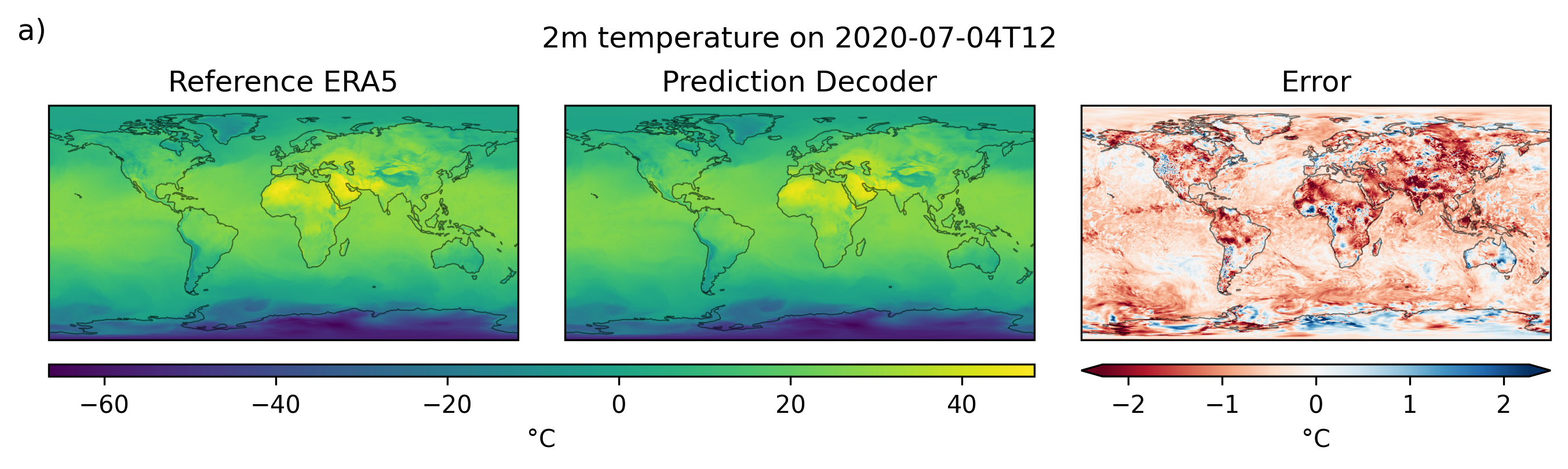}
        \includegraphics[width=0.9\linewidth]{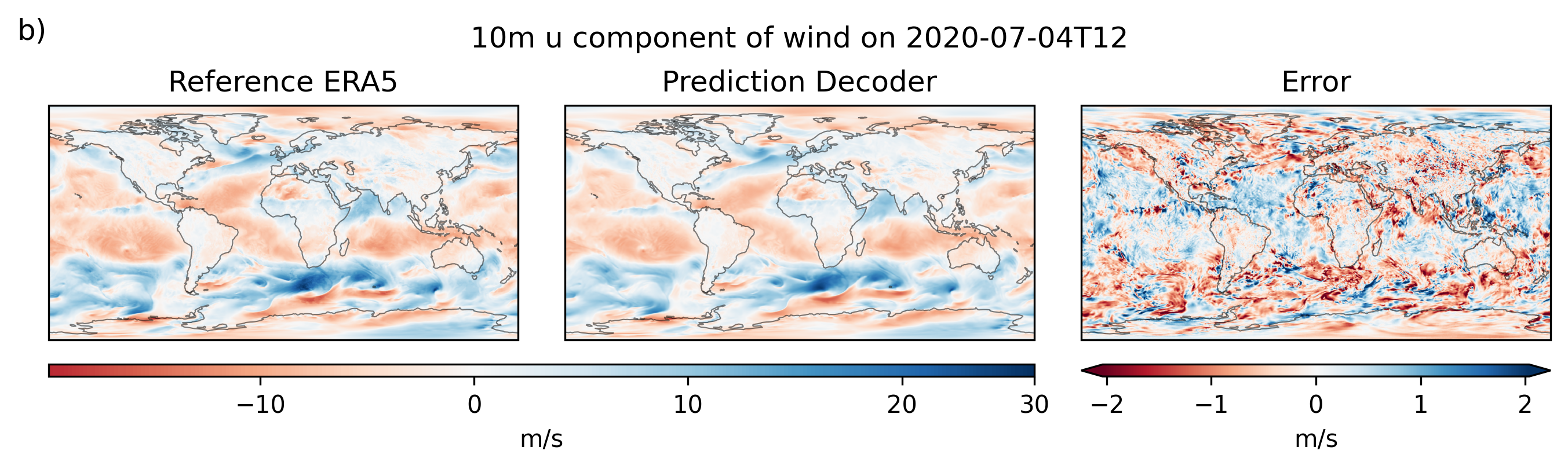}
        \includegraphics[width=0.9\linewidth]{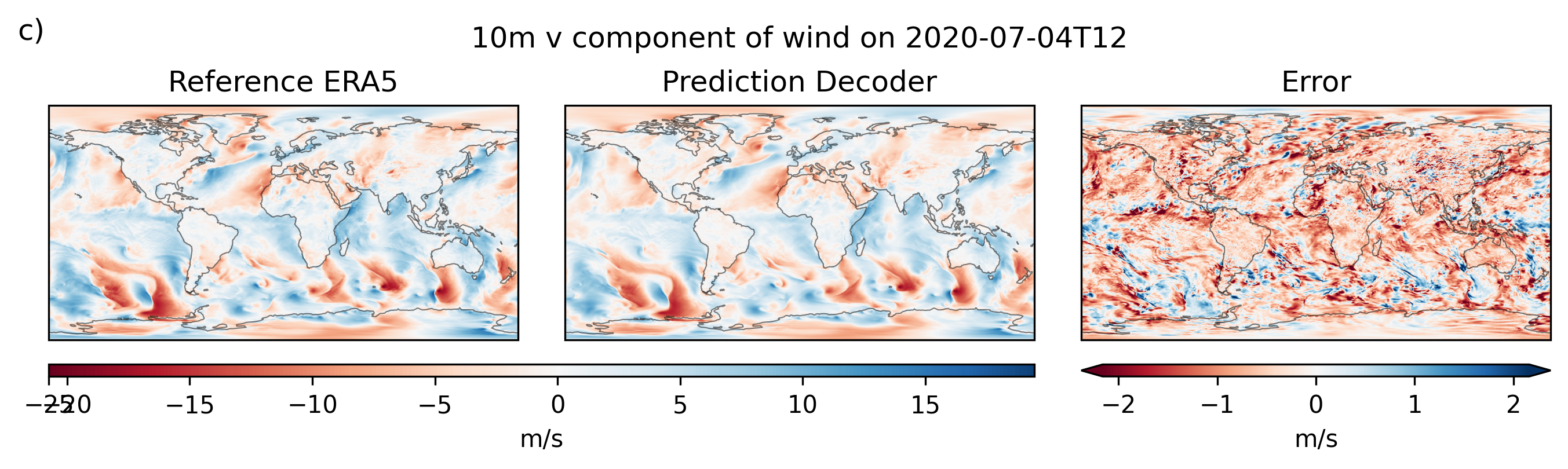}
        \includegraphics[width=0.9\linewidth]{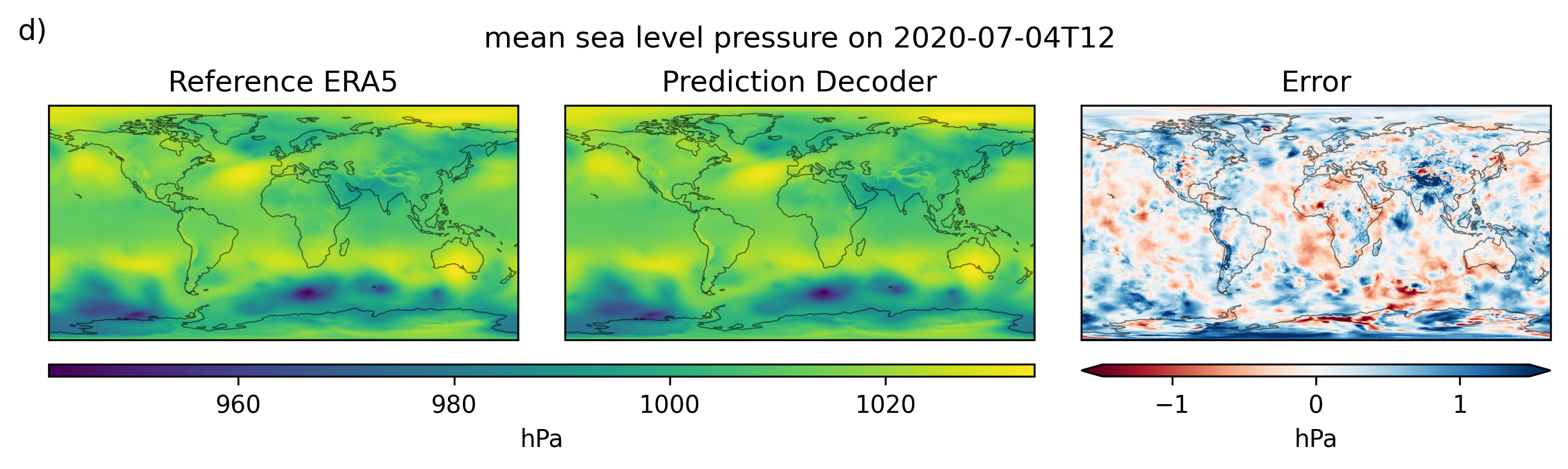}
        \caption{Next step prediction of a) 2-metre temperature, b) 10-metre eastward wind velocity, c) 10-metre northward wind velocity, d) mean sea level pressure. Left column is the reference, middle column is the prediction from frozen Aurora, Right column is the point-wise error}
        \label{fig:maps_original_aurora}
    \end{figure}

\clearpage
\section{Energy spectra}
    \begin{figure}[h]
        \centering
        \includegraphics[width=0.48\linewidth]{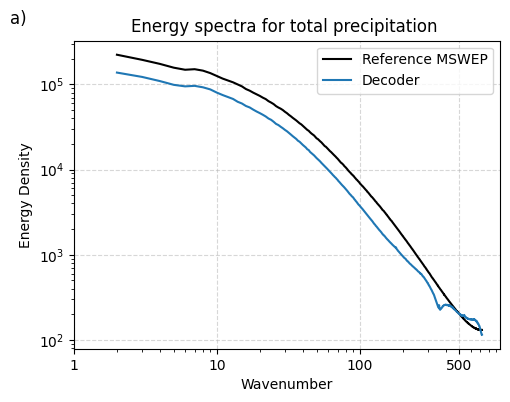}
        \includegraphics[width=0.48\linewidth]{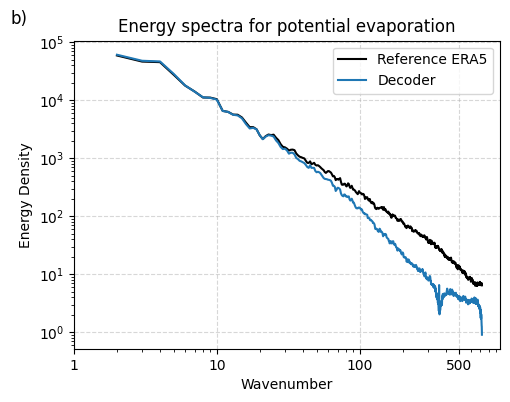}
        \includegraphics[width=0.48\linewidth]{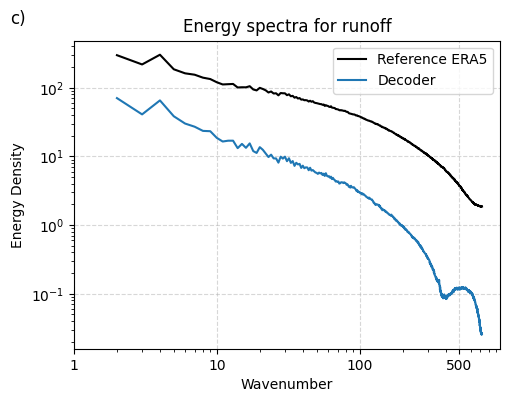}
        \includegraphics[width=0.48\linewidth]{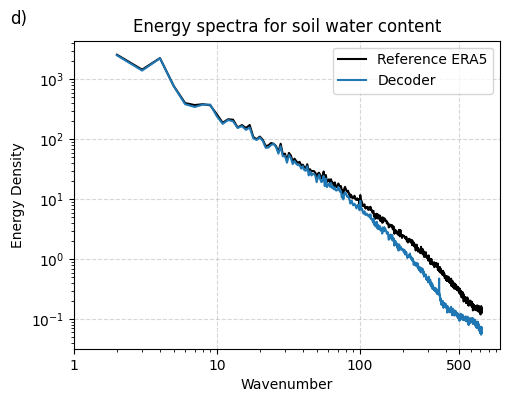}
        \caption{Energy spectra of a) total precipitation, b) potential evaporation, c) runoff, d) soil moisture. Spectra are computed along the longitude dimension for latitude-weighted variables and averaged over the time dimension.}
        \label{fig:energy_spectra}
    \end{figure}

\clearpage
\section{Additional Results Precipitation}
    \begin{table}[h]
    \caption{Precipitation metrics of our decoder (first column) and our Aurora\textsuperscript{+} predictions (second column). Predictions are done with a 6-hour lead time in 2020. The reference dataset is ERA5 instead of MSWEP used in Tab.~\ref{tab:metrics_precip}}
    \label{tab:metrics_precip_era5}
    \centering
    \begin{tabular}{|l|ccccc|}
        \hline
         & Decoder & Aurora\textsuperscript{+} & IFS & GraphCast & FuXi \\
        \hline
        MAE (mm) & 0.388 & 0.306 & 0.379 & 0.220 & \textbf{0.191} \\
        \hline
        RMSE (mm) & 1.693 & 1.341 & 1.629 & \textbf{0.828} & 0.831 \\
        \hline
        SEEPS & 0.400 & 0.297 & 0.288 & 0.234 & \textbf{0.199} \\
        \hline
        W1 & 0.276 & 0.207 & \textbf{0.068} & 0.183 & 0.222 \\
        \hline
        W1 log & 0.038 & 0.026 & \textbf{0.012} & 0.053 & 0.043 \\
        \hline
        bias (mm) & -0.162 & -0.124 & 0.076 & \textbf{-0.046} & -0.049 \\
        \hline
        FSS 1mm & 0.885 & 0.924 & 0.928 & 0.970 & \textbf{0.982} \\
        \hline
        FSS 5mm & 0.725 & 0.846 & 0.854 & 0.941 & \textbf{0.954} \\
        \hline
        PCC & 0.622 & 0.800 & 0.695 & 0.922 & \textbf{0.928} \\
        \hline
        \end{tabular}
    \end{table}

    \begin{figure}[h]
        \centering
        \includegraphics[width=\linewidth]{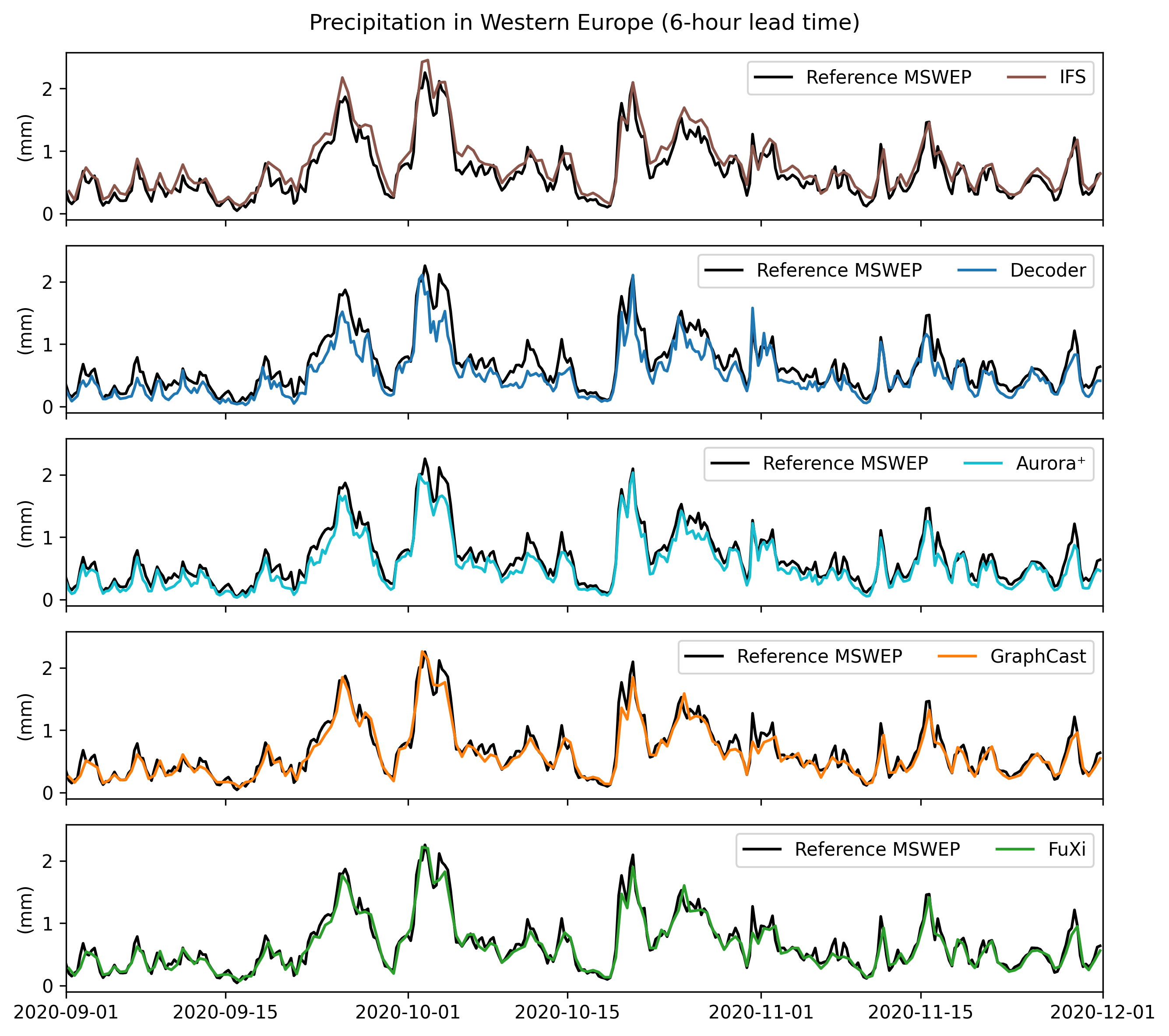}
        \caption{Precipitation prediction during the extratropical cyclone Alex (early October) with a 6-hour lead time from IFS ensemble, our decoder, our Aurora\textsuperscript{+}, GraphCast, FuXi. Time series show the spatial average in a box encompassing Western Europe (i.e., the mean over all points in the red box shown in Fig.~\ref{fig:precip_europe_map}).}
        \label{fig:precip_europe}
    \end{figure}

    \begin{figure}[h]
        \centering
        \includegraphics[width=\linewidth]{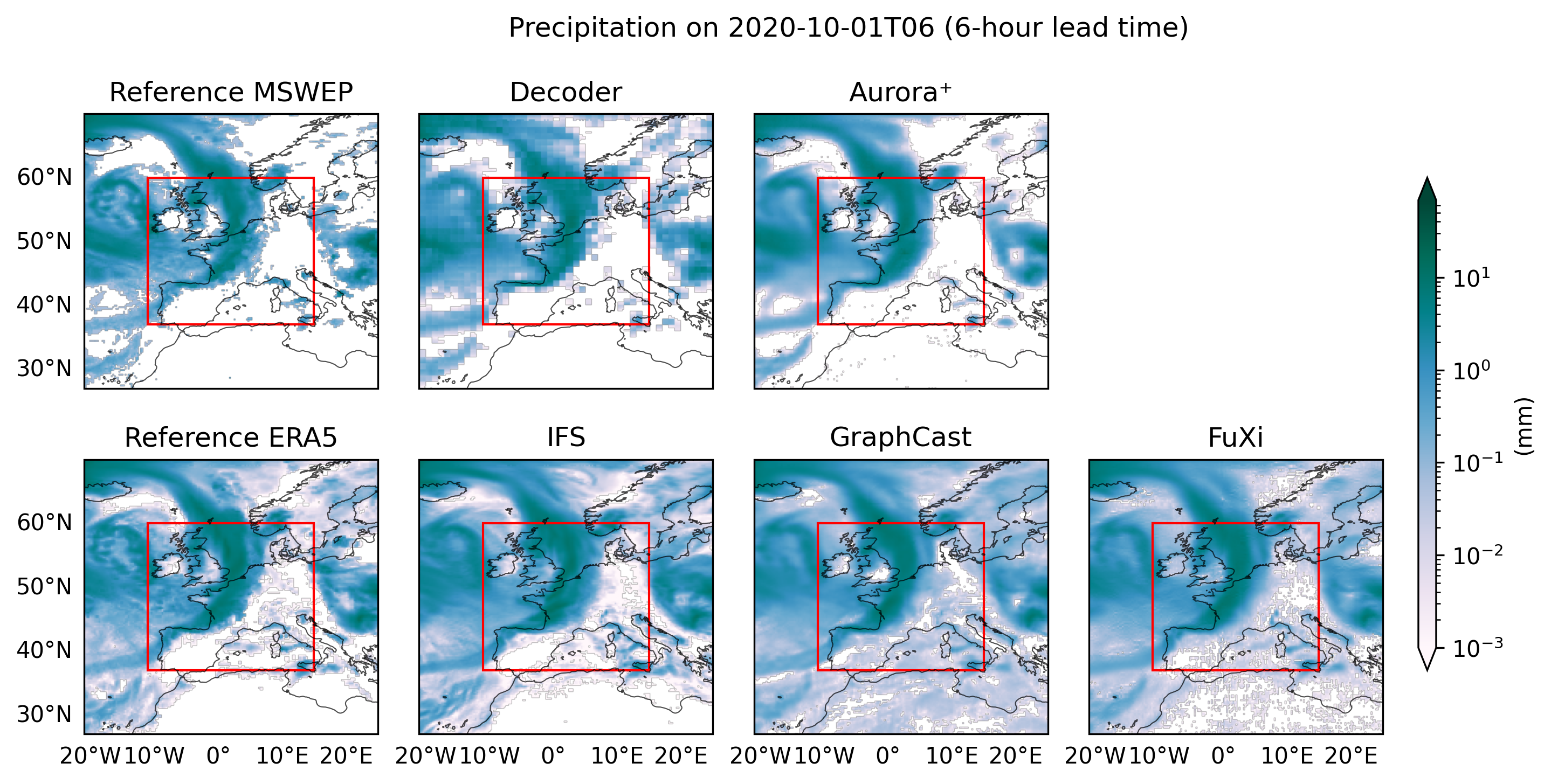}
        \caption{Precipitation in Western Europe on October 1st, 2020, predicted with a 6-hour lead time from IFS ensemble, our decoder, our Aurora\textsuperscript{+}, GraphCast, FuXi. The red box shows the averaging region in Fig.~\ref{fig:precip_europe}.}
        \label{fig:precip_europe_map}
    \end{figure}

\clearpage
\section{Additional Results Temporal Dynamics in River Basins}
    \begin{figure}[h]
        \centering
        \includegraphics[width=\linewidth]{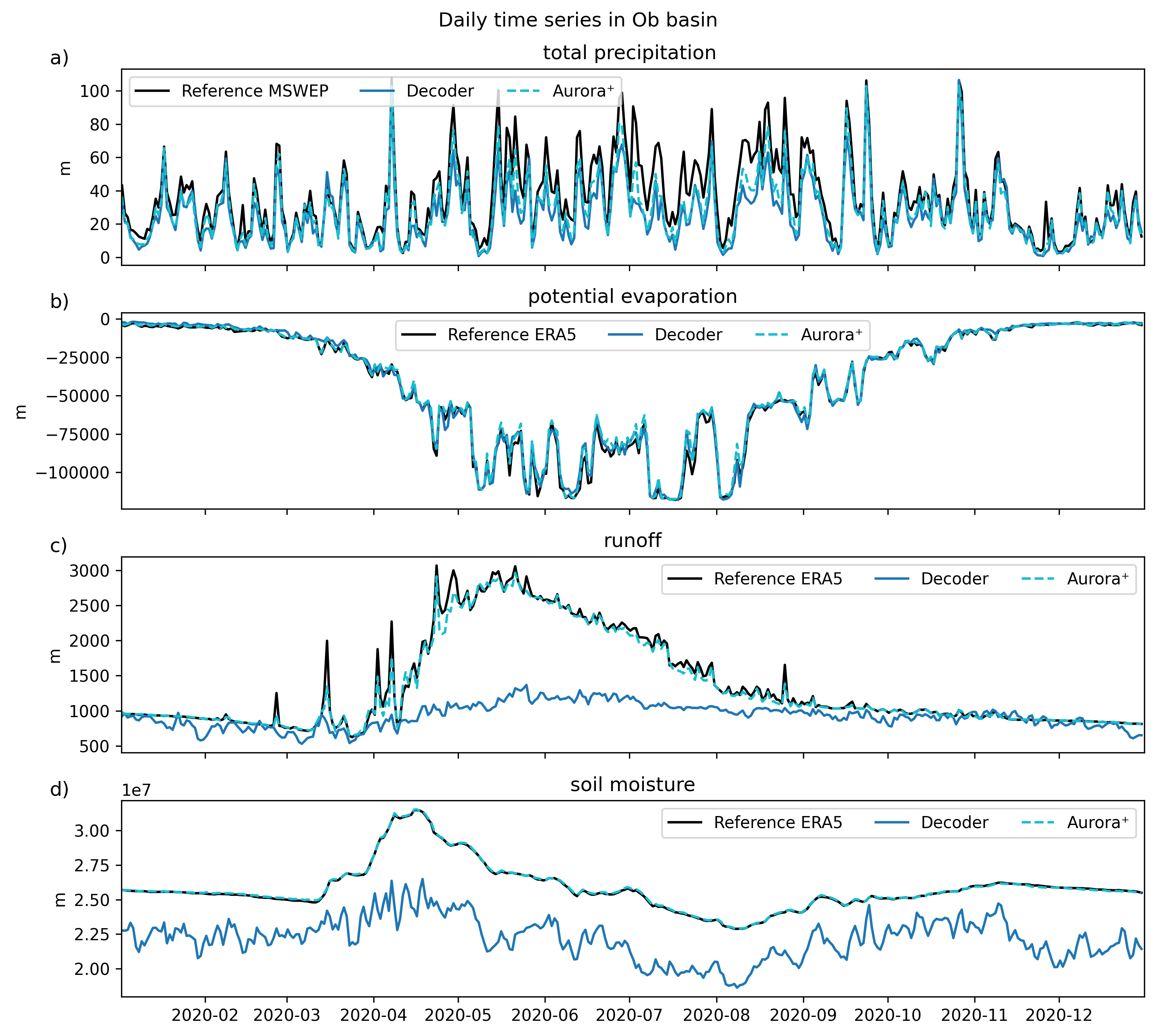}
        \caption{Daily average of 6-hour time series of a) total precipitation [m/6hr], b) potential evaporation [m of equivalent water/6hr], c) runoff [m], d) soil moisture [m] in the Ob basin (shown in green in Fig.~\ref{fig:map_basins}). Time series are obtained as the sum of pixels inside the basin.}
        \label{fig:hydro_basin_cold}
    \end{figure}
    
\clearpage
\section{Regional Plots}
    \begin{figure}[h]
        \centering
        \includegraphics[width=1\linewidth]{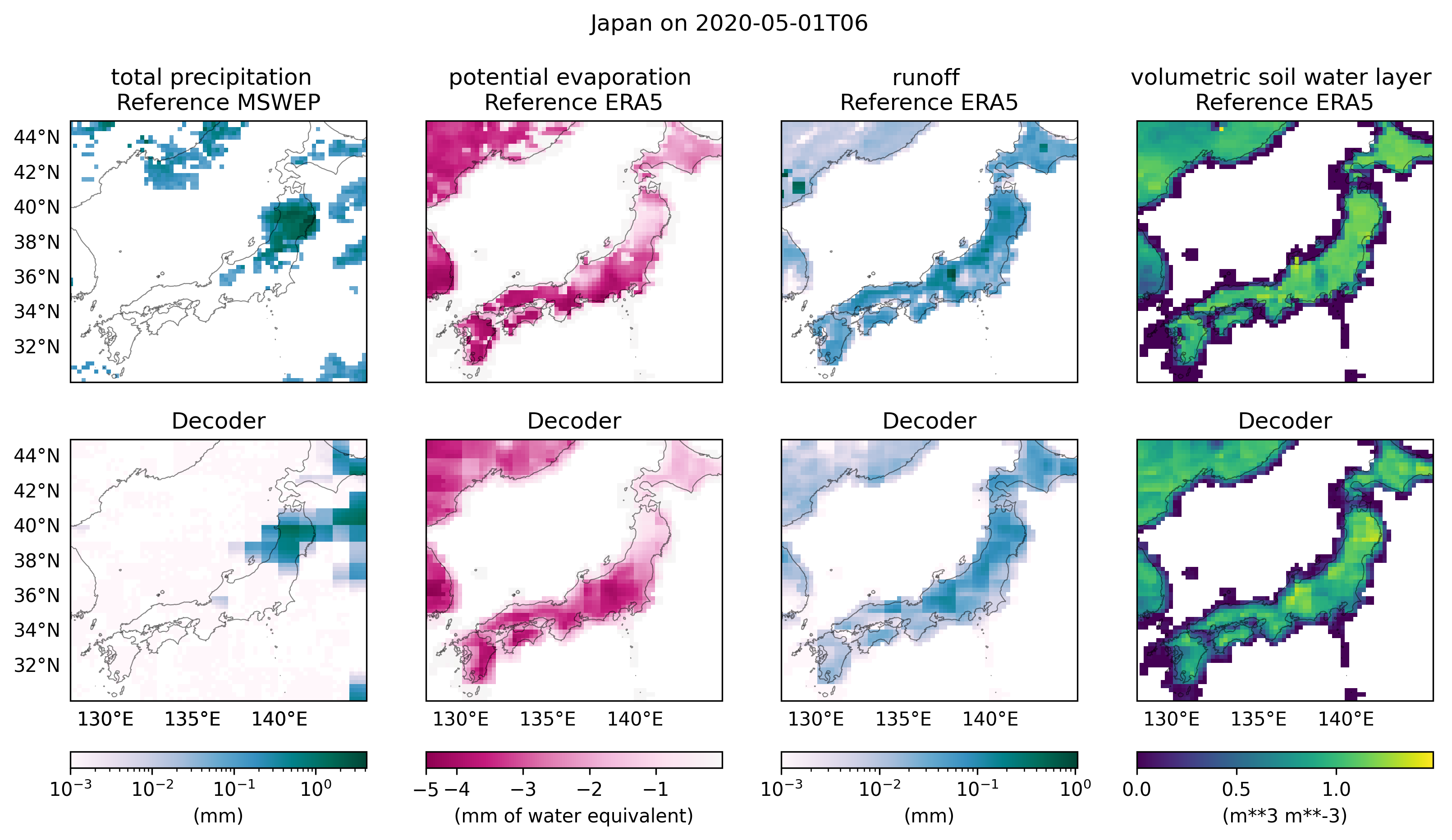}
        \caption{Total precipitation (first column), potential evaporation (second column), runoff (third column), and volumetric soil water layer (fourth column) maps with a zoom around Japan. The decoder prediction (second row) is compared with the reference dataset (first row)}
        \label{fig:maps_highres_japan}
    \end{figure}

\clearpage

\section*{Open Research Section}
    The ERA5 training data and the test data for the benchmark models (IFS, GraphCast, FuXi) were downloaded from the WeatherBench2 Google Cloud bucket \url{https://console.cloud.google.com/storage/browser/weatherbench2}. The MSWEP precipitation dataset is freely available on \url{https://www.gloh2o.org/mswep/}. The inference code of the decoders is available at \url{https://github.com/lehmannfa/aurora-lite-decoder}.

\acknowledgments
The authors sincerely thank Dr.\ Salman Mohebi (SDSC) and Dr.\ Yun Cheng (SDSC) for their contribution to the implementation of the training pipeline. This publication was made possible by an ETH AI Center postdoctoral fellowship to Fanny Lehmann. This work was supported as part of the Swiss AI Initiative by a grant from the Swiss National Supercomputing Centre (CSCS) under project ID a01 on Alps.

\bibliography{references}

\begin{thebibliography}{}

\bibitem [\protect \citeauthoryear {%
Adamov%
\ \protect \BOthers {.}}{%
Adamov%
\ \protect \BOthers {.}}{%
{\protect \APACyear {2025}}%
}]{%
adamov_building_2025}
\APACinsertmetastar {%
adamov_building_2025}%
\begin{APACrefauthors}%
Adamov, S.%
, Oskarsson, J.%
, Denby, L.%
, Landelius, T.%
, Hintz, K.%
, Christiansen, S.%
\BDBL {}Schemm, S.%
\end{APACrefauthors}%
\unskip\
\newblock
\APACrefYearMonthDay{2025}{}{}.
\newblock
\APACrefbtitle {Building {Machine} {Learning} {Limited} {Area} {Models}:
  {Kilometer}-{Scale} {Weather} {Forecasting} in {Realistic} {Settings}.}
  {Building {Machine} {Learning} {Limited} {Area} {Models}: {Kilometer}-{Scale}
  {Weather} {Forecasting} in {Realistic} {Settings}.}
\newblock
\APACaddressPublisher{}{arXiv}.
\newblock
\begin{APACrefDOI} \doi{10.48550/ARXIV.2504.09340} \end{APACrefDOI}
\PrintBackRefs{\CurrentBib}

\bibitem [\protect \citeauthoryear {%
Beck%
\ \protect \BOthers {.}}{%
Beck%
\ \protect \BOthers {.}}{%
{\protect \APACyear {2019}}%
}]{%
beck_mswep_2019}
\APACinsertmetastar {%
beck_mswep_2019}%
\begin{APACrefauthors}%
Beck, H\BPBI E.%
, Wood, E\BPBI F.%
, Pan, M.%
, Fisher, C\BPBI K.%
, Miralles, D\BPBI G.%
, Van~Dijk, A\BPBI I\BPBI J\BPBI M.%
\BDBL {}Adler, R\BPBI F.%
\end{APACrefauthors}%
\unskip\
\newblock
\APACrefYearMonthDay{2019}{}{}.
\newblock
{\BBOQ}\APACrefatitle {{MSWEP} {V2} {Global} 3-{Hourly} 0.1° {Precipitation}:
  {Methodology} and {Quantitative} {Assessment}} {{MSWEP} {V2} {Global}
  3-{Hourly} 0.1° {Precipitation}: {Methodology} and {Quantitative}
  {Assessment}}.{\BBCQ}
\newblock
\APACjournalVolNumPages{Bulletin of the American Meteorological
  Society}{100}{3}{473--500}.
\newblock
\begin{APACrefDOI} \doi{10.1175/BAMS-D-17-0138.1} \end{APACrefDOI}
\PrintBackRefs{\CurrentBib}

\bibitem [\protect \citeauthoryear {%
Bodnar%
\ \protect \BOthers {.}}{%
Bodnar%
\ \protect \BOthers {.}}{%
{\protect \APACyear {2025}}%
}]{%
bodnar_foundation_2025}
\APACinsertmetastar {%
bodnar_foundation_2025}%
\begin{APACrefauthors}%
Bodnar, C.%
, Bruinsma, W\BPBI P.%
, Lucic, A.%
, Stanley, M.%
, Allen, A.%
, Brandstetter, J.%
\BDBL {}Perdikaris, P.%
\end{APACrefauthors}%
\unskip\
\newblock
\APACrefYearMonthDay{2025}{}{}.
\newblock
{\BBOQ}\APACrefatitle {A foundation model for the {Earth} system} {A foundation
  model for the {Earth} system}.{\BBCQ}
\newblock
\APACjournalVolNumPages{Nature}{}{}{}.
\newblock
\begin{APACrefDOI} \doi{10.1038/s41586-025-09005-y} \end{APACrefDOI}
\PrintBackRefs{\CurrentBib}

\bibitem [\protect \citeauthoryear {%
Bommasani%
\ \protect \BOthers {.}}{%
Bommasani%
\ \protect \BOthers {.}}{%
{\protect \APACyear {2021}}%
}]{%
bommasani_opportunities_2021}
\APACinsertmetastar {%
bommasani_opportunities_2021}%
\begin{APACrefauthors}%
Bommasani, R.%
, Hudson, D\BPBI A.%
, Adeli, E.%
, Altman, R.%
, Arora, S.%
, von Arx, S.%
\BDBL {}Liang, P.%
\end{APACrefauthors}%
\unskip\
\newblock
\APACrefYearMonthDay{2021}{}{}.
\newblock
\APACrefbtitle {On the opportunities and risks of foundation models.} {On the
  opportunities and risks of foundation models.}
\newblock
\begin{APACrefURL} \url{https://crfm.stanford.edu/assets/report.pdf}
  \end{APACrefURL}
\PrintBackRefs{\CurrentBib}

\bibitem [\protect \citeauthoryear {%
Chen%
\ \protect \BOthers {.}}{%
Chen%
\ \protect \BOthers {.}}{%
{\protect \APACyear {2023}}%
}]{%
chen_fuxi_2023}
\APACinsertmetastar {%
chen_fuxi_2023}%
\begin{APACrefauthors}%
Chen, L.%
, Zhong, X.%
, Zhang, F.%
, Cheng, Y.%
, Xu, Y.%
, Qi, Y.%
\BCBL {}\ \BBA {} Li, H.%
\end{APACrefauthors}%
\unskip\
\newblock
\APACrefYearMonthDay{2023}{}{}.
\newblock
{\BBOQ}\APACrefatitle {{FuXi}: a cascade machine learning forecasting system
  for 15-day global weather forecast} {{FuXi}: a cascade machine learning
  forecasting system for 15-day global weather forecast}.{\BBCQ}
\newblock
\APACjournalVolNumPages{npj Climate and Atmospheric Science}{6}{1}{190}.
\newblock
\begin{APACrefDOI} \doi{10.1038/s41612-023-00512-1} \end{APACrefDOI}
\PrintBackRefs{\CurrentBib}

\bibitem [\protect \citeauthoryear {%
Gebrechorkos%
\ \protect \BOthers {.}}{%
Gebrechorkos%
\ \protect \BOthers {.}}{%
{\protect \APACyear {2024}}%
}]{%
gebrechorkos_global-scale_2024}
\APACinsertmetastar {%
gebrechorkos_global-scale_2024}%
\begin{APACrefauthors}%
Gebrechorkos, S\BPBI H.%
, Leyland, J.%
, Dadson, S\BPBI J.%
, Cohen, S.%
, Slater, L.%
, Wortmann, M.%
\BDBL {}Darby, S\BPBI E.%
\end{APACrefauthors}%
\unskip\
\newblock
\APACrefYearMonthDay{2024}{}{}.
\newblock
{\BBOQ}\APACrefatitle {Global-scale evaluation of precipitation datasets for
  hydrological modelling} {Global-scale evaluation of precipitation datasets
  for hydrological modelling}.{\BBCQ}
\newblock
\APACjournalVolNumPages{Hydrology and Earth System
  Sciences}{28}{14}{3099--3118}.
\newblock
\begin{APACrefDOI} \doi{10.5194/hess-28-3099-2024} \end{APACrefDOI}
\PrintBackRefs{\CurrentBib}

\bibitem [\protect \citeauthoryear {%
Gou%
\ \BBA {} Soja%
}{%
Gou%
\ \BBA {} Soja%
}{%
{\protect \APACyear {2024}}%
}]{%
gou_global_2024}
\APACinsertmetastar {%
gou_global_2024}%
\begin{APACrefauthors}%
Gou, J.%
\BCBT {}\ \BBA {} Soja, B.%
\end{APACrefauthors}%
\unskip\
\newblock
\APACrefYearMonthDay{2024}{}{}.
\newblock
{\BBOQ}\APACrefatitle {Global high-resolution total water storage anomalies
  from self-supervised data assimilation using deep learning algorithms}
  {Global high-resolution total water storage anomalies from self-supervised
  data assimilation using deep learning algorithms}.{\BBCQ}
\newblock
\APACjournalVolNumPages{Nature Water}{2}{2}{139--150}.
\newblock
\begin{APACrefDOI} \doi{10.1038/s44221-024-00194-w} \end{APACrefDOI}
\PrintBackRefs{\CurrentBib}

\bibitem [\protect \citeauthoryear {%
Hardy%
\ \BBA {} Finney%
}{%
Hardy%
\ \BBA {} Finney%
}{%
{\protect \APACyear {2025}}%
}]{%
hardy_leveraging_2025}
\APACinsertmetastar {%
hardy_leveraging_2025}%
\begin{APACrefauthors}%
Hardy, L.%
\BCBT {}\ \BBA {} Finney, I.%
\end{APACrefauthors}%
\unskip\
\newblock
\APACrefYearMonthDay{2025}{}{}.
\newblock
{\BBOQ}\APACrefatitle {Leveraging state‐of‐the‐art {AI} models to
  forecast wind power generation using deep learning} {Leveraging
  state‐of‐the‐art {AI} models to forecast wind power generation using
  deep learning}.{\BBCQ}
\newblock
\APACjournalVolNumPages{Meteorological Applications}{32}{2}{e70038}.
\newblock
\begin{APACrefDOI} \doi{10.1002/met.70038} \end{APACrefDOI}
\PrintBackRefs{\CurrentBib}

\bibitem [\protect \citeauthoryear {%
Herde%
\ \protect \BOthers {.}}{%
Herde%
\ \protect \BOthers {.}}{%
{\protect \APACyear {2024}}%
}]{%
herde_poseidon_2024}
\APACinsertmetastar {%
herde_poseidon_2024}%
\begin{APACrefauthors}%
Herde, M.%
, Raonić, B.%
, Rohner, T.%
, Käppeli, R.%
, Molinaro, R.%
, de Bézenac, E.%
\BCBL {}\ \BBA {} Mishra, S.%
\end{APACrefauthors}%
\unskip\
\newblock
\APACrefYearMonthDay{2024}{{\APACmonth{05}}}{}.
\newblock
{\BBOQ}\APACrefatitle {Poseidon: {Efficient} {Foundation} {Models} for {PDEs}}
  {Poseidon: {Efficient} {Foundation} {Models} for {PDEs}}.{\BBCQ}
\newblock
\BIn{} \APACrefbtitle {Advances in neural information processing systems.}
  {Advances in neural information processing systems.}
\newblock
\APACaddressPublisher{}{arXiv}.
\newblock
\begin{APACrefURL}
  [{2024-06-02}]\url{https://openreview.net/forum?id=JC1VKK3UXk}
  \end{APACrefURL}
\newblock
\APACrefnote{foundation}
\PrintBackRefs{\CurrentBib}

\bibitem [\protect \citeauthoryear {%
Hersbach%
\ \protect \BOthers {.}}{%
Hersbach%
\ \protect \BOthers {.}}{%
{\protect \APACyear {2020}}%
}]{%
hersbach_era5_2020}
\APACinsertmetastar {%
hersbach_era5_2020}%
\begin{APACrefauthors}%
Hersbach, H.%
, Bell, B.%
, Berrisford, P.%
, Hirahara, S.%
, Horányi, A.%
, Muñoz-Sabater, J.%
\BDBL {}Thépaut, J\BHBI N.%
\end{APACrefauthors}%
\unskip\
\newblock
\APACrefYearMonthDay{2020}{}{}.
\newblock
{\BBOQ}\APACrefatitle {The {ERA5} global reanalysis} {The {ERA5} global
  reanalysis}.{\BBCQ}
\newblock
\APACjournalVolNumPages{Quarterly Journal of the Royal Meteorological
  Society}{146}{730}{1999--2049}.
\newblock
\begin{APACrefDOI} \doi{10.1002/qj.3803} \end{APACrefDOI}
\PrintBackRefs{\CurrentBib}

\bibitem [\protect \citeauthoryear {%
Jaegle%
\ \protect \BOthers {.}}{%
Jaegle%
\ \protect \BOthers {.}}{%
{\protect \APACyear {2021}}%
}]{%
jaegle_perceiver_2021}
\APACinsertmetastar {%
jaegle_perceiver_2021}%
\begin{APACrefauthors}%
Jaegle, A.%
, Gimeno, F.%
, Brock, A.%
, Vinyals, O.%
, Zisserman, A.%
\BCBL {}\ \BBA {} Carreira, J.%
\end{APACrefauthors}%
\unskip\
\newblock
\APACrefYearMonthDay{2021}{}{}.
\newblock
{\BBOQ}\APACrefatitle {Perceiver: {General} {Perception} with {Iterative}
  {Attention}} {Perceiver: {General} {Perception} with {Iterative}
  {Attention}}.{\BBCQ}
\newblock
\BIn{} M.~Meila\ \BBA {} T.~Zhang\ (\BEDS), \APACrefbtitle {Proceedings of the
  38th {International} {Conference} on {Machine} {Learning}} {Proceedings of
  the 38th {International} {Conference} on {Machine} {Learning}}\ (\BVOL~139,
  \BPGS\ 4651--4664).
\newblock
\APACaddressPublisher{}{PMLR}.
\newblock
\begin{APACrefURL} \url{https://proceedings.mlr.press/v139/jaegle21a.html}
  \end{APACrefURL}
\PrintBackRefs{\CurrentBib}

\bibitem [\protect \citeauthoryear {%
Lam%
\ \protect \BOthers {.}}{%
Lam%
\ \protect \BOthers {.}}{%
{\protect \APACyear {2023}}%
}]{%
lam_learning_2023}
\APACinsertmetastar {%
lam_learning_2023}%
\begin{APACrefauthors}%
Lam, R.%
, Sanchez-Gonzalez, A.%
, Willson, M.%
, Wirnsberger, P.%
, Fortunato, M.%
, Alet, F.%
\BDBL {}Battaglia, P.%
\end{APACrefauthors}%
\unskip\
\newblock
\APACrefYearMonthDay{2023}{}{}.
\newblock
{\BBOQ}\APACrefatitle {Learning skillful medium-range global weather
  forecasting} {Learning skillful medium-range global weather
  forecasting}.{\BBCQ}
\newblock
\APACjournalVolNumPages{Science}{}{}{eadi2336}.
\newblock
\begin{APACrefDOI} \doi{10.1126/science.adi2336} \end{APACrefDOI}
\PrintBackRefs{\CurrentBib}

\bibitem [\protect \citeauthoryear {%
Lehner%
\ \BBA {} Grill%
}{%
Lehner%
\ \BBA {} Grill%
}{%
{\protect \APACyear {2013}}%
}]{%
lehner_global_2013}
\APACinsertmetastar {%
lehner_global_2013}%
\begin{APACrefauthors}%
Lehner, B.%
\BCBT {}\ \BBA {} Grill, G.%
\end{APACrefauthors}%
\unskip\
\newblock
\APACrefYearMonthDay{2013}{}{}.
\newblock
{\BBOQ}\APACrefatitle {Global river hydrography and network routing: baseline
  data and new approaches to study the world's large river systems} {Global
  river hydrography and network routing: baseline data and new approaches to
  study the world's large river systems}.{\BBCQ}
\newblock
\APACjournalVolNumPages{Hydrological Processes}{27}{15}{2171--2186}.
\newblock
\begin{APACrefDOI} \doi{10.1002/hyp.9740} \end{APACrefDOI}
\PrintBackRefs{\CurrentBib}

\bibitem [\protect \citeauthoryear {%
Lessig%
\ \protect \BOthers {.}}{%
Lessig%
\ \protect \BOthers {.}}{%
{\protect \APACyear {2023}}%
}]{%
lessig_atmorep_2023}
\APACinsertmetastar {%
lessig_atmorep_2023}%
\begin{APACrefauthors}%
Lessig, C.%
, Luise, I.%
, Gong, B.%
, Langguth, M.%
, Stadtler, S.%
\BCBL {}\ \BBA {} Schultz, M.%
\end{APACrefauthors}%
\unskip\
\newblock
\APACrefYearMonthDay{2023}{}{}.
\newblock
\APACrefbtitle {{AtmoRep}: {A} stochastic model of atmosphere dynamics using
  large scale representation learning.} {{AtmoRep}: {A} stochastic model of
  atmosphere dynamics using large scale representation learning.}
\newblock
\APACaddressPublisher{}{arXiv}.
\newblock
\begin{APACrefDOI} \doi{10.48550/ARXIV.2308.13280} \end{APACrefDOI}
\PrintBackRefs{\CurrentBib}

\bibitem [\protect \citeauthoryear {%
Li%
\ \protect \BOthers {.}}{%
Li%
\ \protect \BOthers {.}}{%
{\protect \APACyear {2024}}%
}]{%
li_forecasting_2024}
\APACinsertmetastar {%
li_forecasting_2024}%
\begin{APACrefauthors}%
Li, F.%
, Kusche, J.%
, Sneeuw, N.%
, Siebert, S.%
, Gerdener, H.%
, Wang, Z.%
\BDBL {}Tian, K.%
\end{APACrefauthors}%
\unskip\
\newblock
\APACrefYearMonthDay{2024}{}{}.
\newblock
{\BBOQ}\APACrefatitle {Forecasting {Next} {Year}'s {Global} {Land} {Water}
  {Storage} {Using} {GRACE} {Data}} {Forecasting {Next} {Year}'s {Global}
  {Land} {Water} {Storage} {Using} {GRACE} {Data}}.{\BBCQ}
\newblock
\APACjournalVolNumPages{Geophysical Research Letters}{51}{17}{e2024GL109101}.
\newblock
\begin{APACrefDOI} \doi{10.1029/2024GL109101} \end{APACrefDOI}
\PrintBackRefs{\CurrentBib}

\bibitem [\protect \citeauthoryear {%
Liu%
\ \protect \BOthers {.}}{%
Liu%
\ \protect \BOthers {.}}{%
{\protect \APACyear {2024}}%
}]{%
liu_evaluation_2024}
\APACinsertmetastar {%
liu_evaluation_2024}%
\begin{APACrefauthors}%
Liu, C\BHBI C.%
, Hsu, K.%
, Peng, M\BPBI S.%
, Chen, D\BHBI S.%
, Chang, P\BHBI L.%
, Hsiao, L\BHBI F.%
\BDBL {}Kuo, H\BHBI C.%
\end{APACrefauthors}%
\unskip\
\newblock
\APACrefYearMonthDay{2024}{}{}.
\newblock
{\BBOQ}\APACrefatitle {Evaluation of five global {AI} models for predicting
  weather in {Eastern} {Asia} and {Western} {Pacific}} {Evaluation of five
  global {AI} models for predicting weather in {Eastern} {Asia} and {Western}
  {Pacific}}.{\BBCQ}
\newblock
\APACjournalVolNumPages{npj Climate and Atmospheric Science}{7}{1}{221}.
\newblock
\begin{APACrefDOI} \doi{10.1038/s41612-024-00769-0} \end{APACrefDOI}
\PrintBackRefs{\CurrentBib}

\bibitem [\protect \citeauthoryear {%
Nguyen%
, Brandstetter%
, Kapoor%
, Gupta%
\BCBL {}\ \BBA {} Grover%
}{%
Nguyen%
\ \protect \BOthers {.}}{%
{\protect \APACyear {2023}}%
}]{%
nguyen_climax_2023}
\APACinsertmetastar {%
nguyen_climax_2023}%
\begin{APACrefauthors}%
Nguyen, T.%
, Brandstetter, J.%
, Kapoor, A.%
, Gupta, J\BPBI K.%
\BCBL {}\ \BBA {} Grover, A.%
\end{APACrefauthors}%
\unskip\
\newblock
\APACrefYearMonthDay{2023}{}{}.
\newblock
{\BBOQ}\APACrefatitle {{ClimaX}: {A} foundation model for weather and climate}
  {{ClimaX}: {A} foundation model for weather and climate}.{\BBCQ}
\newblock
\BIn{} A.~Krause, E.~Brunskill, K.~Cho, B.~Engelhardt, S.~Sabato\BCBL {}\ \BBA
  {} J.~Scarlett\ (\BEDS), \APACrefbtitle {Proceedings of the 40th
  international conference on machine learning} {Proceedings of the 40th
  international conference on machine learning}\ (\BVOL~202, \BPGS\
  25904--25938).
\newblock
\begin{APACrefURL} \url{https://proceedings.mlr.press/v202/nguyen23a.html}
  \end{APACrefURL}
\PrintBackRefs{\CurrentBib}

\bibitem [\protect \citeauthoryear {%
Palazzoli%
, Ceola%
\BCBL {}\ \BBA {} Gentine%
}{%
Palazzoli%
\ \protect \BOthers {.}}{%
{\protect \APACyear {2025}}%
}]{%
palazzoli_graice_2025}
\APACinsertmetastar {%
palazzoli_graice_2025}%
\begin{APACrefauthors}%
Palazzoli, I.%
, Ceola, S.%
\BCBL {}\ \BBA {} Gentine, P.%
\end{APACrefauthors}%
\unskip\
\newblock
\APACrefYearMonthDay{2025}{}{}.
\newblock
{\BBOQ}\APACrefatitle {{GRAiCE}: reconstructing terrestrial water storage
  anomalies with recurrent neural networks} {{GRAiCE}: reconstructing
  terrestrial water storage anomalies with recurrent neural networks}.{\BBCQ}
\newblock
\APACjournalVolNumPages{Scientific Data}{12}{1}{146}.
\newblock
\begin{APACrefDOI} \doi{10.1038/s41597-025-04403-3} \end{APACrefDOI}
\PrintBackRefs{\CurrentBib}

\bibitem [\protect \citeauthoryear {%
Ramdas%
, Garcia%
\BCBL {}\ \BBA {} Cuturi%
}{%
Ramdas%
\ \protect \BOthers {.}}{%
{\protect \APACyear {2015}}%
}]{%
ramdas_wasserstein_2015}
\APACinsertmetastar {%
ramdas_wasserstein_2015}%
\begin{APACrefauthors}%
Ramdas, A.%
, Garcia, N.%
\BCBL {}\ \BBA {} Cuturi, M.%
\end{APACrefauthors}%
\unskip\
\newblock
\APACrefYearMonthDay{2015}{}{}.
\newblock
\APACrefbtitle {On {Wasserstein} {Two} {Sample} {Testing} and {Related}
  {Families} of {Nonparametric} {Tests}.} {On {Wasserstein} {Two} {Sample}
  {Testing} and {Related} {Families} of {Nonparametric} {Tests}.}
\newblock
\APACaddressPublisher{}{arXiv}.
\newblock
\begin{APACrefDOI} \doi{10.48550/ARXIV.1509.02237} \end{APACrefDOI}
\PrintBackRefs{\CurrentBib}

\bibitem [\protect \citeauthoryear {%
Rasp%
\ \protect \BOthers {.}}{%
Rasp%
\ \protect \BOthers {.}}{%
{\protect \APACyear {2020}}%
}]{%
rasp_weatherbench_2020}
\APACinsertmetastar {%
rasp_weatherbench_2020}%
\begin{APACrefauthors}%
Rasp, S.%
, Dueben, P\BPBI D.%
, Scher, S.%
, Weyn, J\BPBI A.%
, Mouatadid, S.%
\BCBL {}\ \BBA {} Thuerey, N.%
\end{APACrefauthors}%
\unskip\
\newblock
\APACrefYearMonthDay{2020}{}{}.
\newblock
{\BBOQ}\APACrefatitle {{WeatherBench}: {A} {Benchmark} {Data} {Set} for
  {Data}‐{Driven} {Weather} {Forecasting}} {{WeatherBench}: {A} {Benchmark}
  {Data} {Set} for {Data}‐{Driven} {Weather} {Forecasting}}.{\BBCQ}
\newblock
\APACjournalVolNumPages{Journal of Advances in Modeling Earth
  Systems}{12}{11}{e2020MS002203}.
\newblock
\begin{APACrefDOI} \doi{10.1029/2020MS002203} \end{APACrefDOI}
\PrintBackRefs{\CurrentBib}

\bibitem [\protect \citeauthoryear {%
Rasp%
\ \protect \BOthers {.}}{%
Rasp%
\ \protect \BOthers {.}}{%
{\protect \APACyear {2024}}%
}]{%
rasp_weatherbench_2024}
\APACinsertmetastar {%
rasp_weatherbench_2024}%
\begin{APACrefauthors}%
Rasp, S.%
, Hoyer, S.%
, Merose, A.%
, Langmore, I.%
, Battaglia, P.%
, Russel, T.%
\BDBL {}Sha, F.%
\end{APACrefauthors}%
\unskip\
\newblock
\APACrefYearMonthDay{2024}{}{}.
\newblock
\APACrefbtitle {{WeatherBench} 2: {A} benchmark for the next generation of
  data-driven global weather models.} {{WeatherBench} 2: {A} benchmark for the
  next generation of data-driven global weather models.}
\newblock
\APACaddressPublisher{}{arXiv}.
\newblock
\begin{APACrefURL} \url{http://arxiv.org/abs/2308.15560} \end{APACrefURL}
\PrintBackRefs{\CurrentBib}

\bibitem [\protect \citeauthoryear {%
Roberts%
\ \BBA {} Lean%
}{%
Roberts%
\ \BBA {} Lean%
}{%
{\protect \APACyear {2008}}%
}]{%
roberts_scale-selective_2008}
\APACinsertmetastar {%
roberts_scale-selective_2008}%
\begin{APACrefauthors}%
Roberts, N\BPBI M.%
\BCBT {}\ \BBA {} Lean, H\BPBI W.%
\end{APACrefauthors}%
\unskip\
\newblock
\APACrefYearMonthDay{2008}{}{}.
\newblock
{\BBOQ}\APACrefatitle {Scale-{Selective} {Verification} of {Rainfall}
  {Accumulations} from {High}-{Resolution} {Forecasts} of {Convective}
  {Events}} {Scale-{Selective} {Verification} of {Rainfall} {Accumulations}
  from {High}-{Resolution} {Forecasts} of {Convective} {Events}}.{\BBCQ}
\newblock
\APACjournalVolNumPages{Monthly Weather Review}{136}{1}{78--97}.
\newblock
\begin{APACrefDOI} \doi{10.1175/2007MWR2123.1} \end{APACrefDOI}
\PrintBackRefs{\CurrentBib}

\bibitem [\protect \citeauthoryear {%
Rodwell%
, Richardson%
, Hewson%
\BCBL {}\ \BBA {} Haiden%
}{%
Rodwell%
\ \protect \BOthers {.}}{%
{\protect \APACyear {2010}}%
}]{%
rodwell_new_2010}
\APACinsertmetastar {%
rodwell_new_2010}%
\begin{APACrefauthors}%
Rodwell, M\BPBI J.%
, Richardson, D\BPBI S.%
, Hewson, T\BPBI D.%
\BCBL {}\ \BBA {} Haiden, T.%
\end{APACrefauthors}%
\unskip\
\newblock
\APACrefYearMonthDay{2010}{}{}.
\newblock
{\BBOQ}\APACrefatitle {A new equitable score suitable for verifying
  precipitation in numerical weather prediction} {A new equitable score
  suitable for verifying precipitation in numerical weather prediction}.{\BBCQ}
\newblock
\APACjournalVolNumPages{Quarterly Journal of the Royal Meteorological
  Society}{136}{650}{1344--1363}.
\newblock
\begin{APACrefDOI} \doi{10.1002/qj.656} \end{APACrefDOI}
\PrintBackRefs{\CurrentBib}

\bibitem [\protect \citeauthoryear {%
Schmude%
\ \protect \BOthers {.}}{%
Schmude%
\ \protect \BOthers {.}}{%
{\protect \APACyear {2024}}%
}]{%
schmude_prithvi_2024}
\APACinsertmetastar {%
schmude_prithvi_2024}%
\begin{APACrefauthors}%
Schmude, J.%
, Roy, S.%
, Trojak, W.%
, Jakubik, J.%
, Civitarese, D\BPBI S.%
, Singh, S.%
\BDBL {}Ramachandran, R.%
\end{APACrefauthors}%
\unskip\
\newblock
\APACrefYearMonthDay{2024}{}{}.
\newblock
\APACrefbtitle {Prithvi {WxC}: {Foundation} {Model} for {Weather} and
  {Climate}.} {Prithvi {WxC}: {Foundation} {Model} for {Weather} and
  {Climate}.}
\newblock
\APACaddressPublisher{}{arXiv}.
\newblock
\begin{APACrefDOI} \doi{10.48550/ARXIV.2409.13598} \end{APACrefDOI}
\PrintBackRefs{\CurrentBib}

\bibitem [\protect \citeauthoryear {%
Wang%
\ \protect \BOthers {.}}{%
Wang%
\ \protect \BOthers {.}}{%
{\protect \APACyear {2024}}%
}]{%
wang_orbit_2024}
\APACinsertmetastar {%
wang_orbit_2024}%
\begin{APACrefauthors}%
Wang, X.%
, Liu, S.%
, Tsaris, A.%
, Choi, J\BHBI Y.%
, Aji, A\BPBI M.%
, Fan, M.%
\BDBL {}Balaprakash, P.%
\end{APACrefauthors}%
\unskip\
\newblock
\APACrefYearMonthDay{2024}{}{}.
\newblock
{\BBOQ}\APACrefatitle {{ORBIT}: {Oak} ridge base foundation model for earth
  system predictability} {{ORBIT}: {Oak} ridge base foundation model for earth
  system predictability}.{\BBCQ}
\newblock
\BIn{} \APACrefbtitle {{SC24}: {International} conference for high performance
  computing, networking, storage and analysis} {{SC24}: {International}
  conference for high performance computing, networking, storage and analysis}\
  (\BPGS\ 1--11).
\newblock
\begin{APACrefDOI} \doi{10.1109/SC41406.2024.00007} \end{APACrefDOI}
\PrintBackRefs{\CurrentBib}

\bibitem [\protect \citeauthoryear {%
Watt-Meyer%
\ \protect \BOthers {.}}{%
Watt-Meyer%
\ \protect \BOthers {.}}{%
{\protect \APACyear {2024}}%
}]{%
watt-meyer_ace2_2024}
\APACinsertmetastar {%
watt-meyer_ace2_2024}%
\begin{APACrefauthors}%
Watt-Meyer, O.%
, Henn, B.%
, McGibbon, J.%
, Clark, S\BPBI K.%
, Kwa, A.%
, Perkins, W\BPBI A.%
\BDBL {}Bretherton, C\BPBI S.%
\end{APACrefauthors}%
\unskip\
\newblock
\APACrefYearMonthDay{2024}{}{}.
\newblock
\APACrefbtitle {{ACE2}: {Accurately} learning subseasonal to decadal
  atmospheric variability and forced responses.} {{ACE2}: {Accurately} learning
  subseasonal to decadal atmospheric variability and forced responses.}
\newblock
\APACaddressPublisher{}{arXiv}.
\newblock
\begin{APACrefDOI} \doi{10.48550/arXiv.2411.11268} \end{APACrefDOI}
\PrintBackRefs{\CurrentBib}

\bibitem [\protect \citeauthoryear {%
Xu%
\ \protect \BOthers {.}}{%
Xu%
\ \protect \BOthers {.}}{%
{\protect \APACyear {2025}}%
}]{%
xu_artificial_2025}
\APACinsertmetastar {%
xu_artificial_2025}%
\begin{APACrefauthors}%
Xu, P.%
, Zheng, X.%
, Gao, T.%
, Wang, Y.%
, Yin, J.%
, Zhang, J.%
\BDBL {}Chen, X.%
\end{APACrefauthors}%
\unskip\
\newblock
\APACrefYearMonthDay{2025}{}{}.
\newblock
{\BBOQ}\APACrefatitle {An artificial intelligence-based limited area model for
  forecasting of surface meteorological variables} {An artificial
  intelligence-based limited area model for forecasting of surface
  meteorological variables}.{\BBCQ}
\newblock
\APACjournalVolNumPages{Communications Earth \& Environment}{6}{1}{}.
\newblock
\begin{APACrefDOI} \doi{10.1038/s43247-025-02347-5} \end{APACrefDOI}
\PrintBackRefs{\CurrentBib}

\bibitem [\protect \citeauthoryear {%
Zhao%
\ \protect \BOthers {.}}{%
Zhao%
\ \protect \BOthers {.}}{%
{\protect \APACyear {2024}}%
}]{%
zhao_weathergfm_2024}
\APACinsertmetastar {%
zhao_weathergfm_2024}%
\begin{APACrefauthors}%
Zhao, X.%
, Zhou, Z.%
, Zhang, W.%
, Liu, Y.%
, Chen, X.%
, Gong, J.%
\BDBL {}Bai, L.%
\end{APACrefauthors}%
\unskip\
\newblock
\APACrefYearMonthDay{2024}{}{}.
\newblock
\APACrefbtitle {{WeatherGFM}: {Learning} {A} {Weather} {Generalist}
  {Foundation} {Model} via {In}-context {Learning}.} {{WeatherGFM}: {Learning}
  {A} {Weather} {Generalist} {Foundation} {Model} via {In}-context {Learning}.}
\newblock
\APACaddressPublisher{}{arXiv}.
\newblock
\begin{APACrefDOI} \doi{10.48550/ARXIV.2411.05420} \end{APACrefDOI}
\PrintBackRefs{\CurrentBib}

\end{thebibliography}

\end{document}